\documentclass[10pt,twocolumn,letterpaper]{article}

\usepackage[pagenumbers]{cvpr} 

\usepackage{upgreek}
\usepackage{multirow}
\usepackage{multicol}
\usepackage{float}
\usepackage{amssymb}
\usepackage{amsthm}
\usepackage[utf8]{inputenc} 
\usepackage[T1]{fontenc}    
\usepackage{url}   
\usepackage{bbm}
\usepackage{bm}
\usepackage{soul}
\usepackage{booktabs}       
\usepackage{amsfonts}       
\usepackage{nicefrac} 
\usepackage{amsmath}
\usepackage{graphicx}
\usepackage{microtype}      
\usepackage[ruled,vlined]{algorithm2e}
\DeclareMathOperator*{\argmax}{arg\,max}
\usepackage{algpseudocode}

\usepackage{wrapfig}

\usepackage[pagebackref,breaklinks,colorlinks]{hyperref}

\usepackage[capitalize]{cleveref}
\crefname{section}{Sec.}{Secs.}
\Crefname{section}{Section}{Sections}
\Crefname{table}{Table}{Tables}
\crefname{table}{Tab.}{Tabs.}

\def\confName{CVPR}
\def\confYear{2022}

\begin{document}

\title{ExCon: Explanation-driven Supervised Contrastive Learning for Image Classification }

\author{Zhibo Zhang$^1$, Jongseong Jang$^2$, Chiheb Trabelsi$^1$, Ruiwen Li$^1$, Scott Sanner$^1$, Yeonjeong Jeong$^2$\\ Dongsub Shim$^2$\\
$^1$ University of Toronto and $^2$ LG AI Research\\
{\tt\small zhibozhang@cs.toronto.edu, j.jang@lgresearch.ai,chiheb.trabelsi@utoronto.ca}\\{\tt\small ruiwen.li@mail.utoronto.ca,  ssanner@mie.utoronto.ca, yj.jeong@lgresearch.ai} \\{\tt\small dongsub.shim@lgresearch.ai}
}
\maketitle

\begin{abstract}
Contrastive learning has led to substantial improvements in the quality of learned embedding representations for tasks such as image classification. However, a key drawback of existing contrastive augmentation methods is that they may lead to the modification of the image content which can yield undesired alterations of its semantics. This can affect the performance of the model on downstream tasks. Hence, in this paper, we ask whether we can augment image data in contrastive learning such that the task-relevant semantic content of an image is preserved. For this purpose, we propose to leverage saliency-based explanation methods to create content-preserving masked augmentations for contrastive learning. Our novel explanation-driven supervised contrastive learning (ExCon) methodology critically serves the dual goals of encouraging nearby image embeddings to have similar content and explanation. To quantify the impact of ExCon, we conduct experiments on the CIFAR-100 and the Tiny ImageNet datasets. We demonstrate that ExCon outperforms vanilla supervised contrastive learning in terms of classification, explanation quality, adversarial robustness as well as probabilistic calibration in the context of distributional shift. Our code is publicly available at \href{https://github.com/DarrenZhang01/ExCon}{https://github.com/DarrenZhang01/ExCon}.
\end{abstract}

\section{Introduction}

\begin{figure*}

    \centering
        \includegraphics[width=0.49\textwidth]{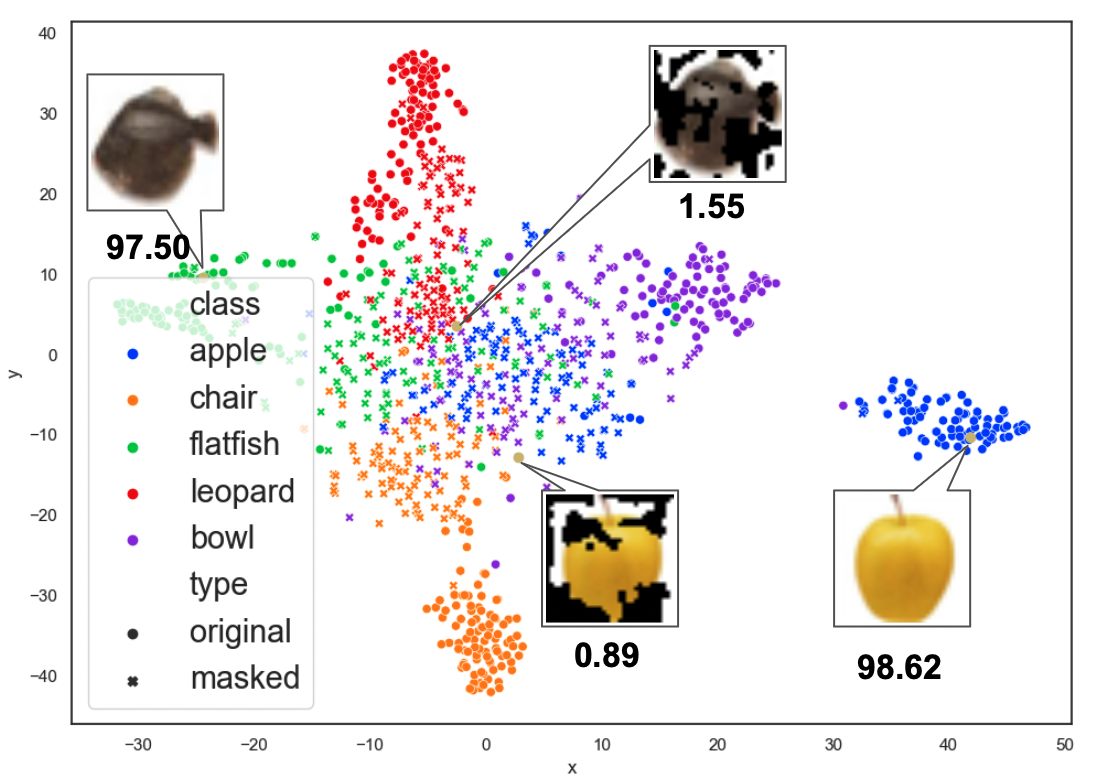}
        \includegraphics[width=0.49\textwidth]{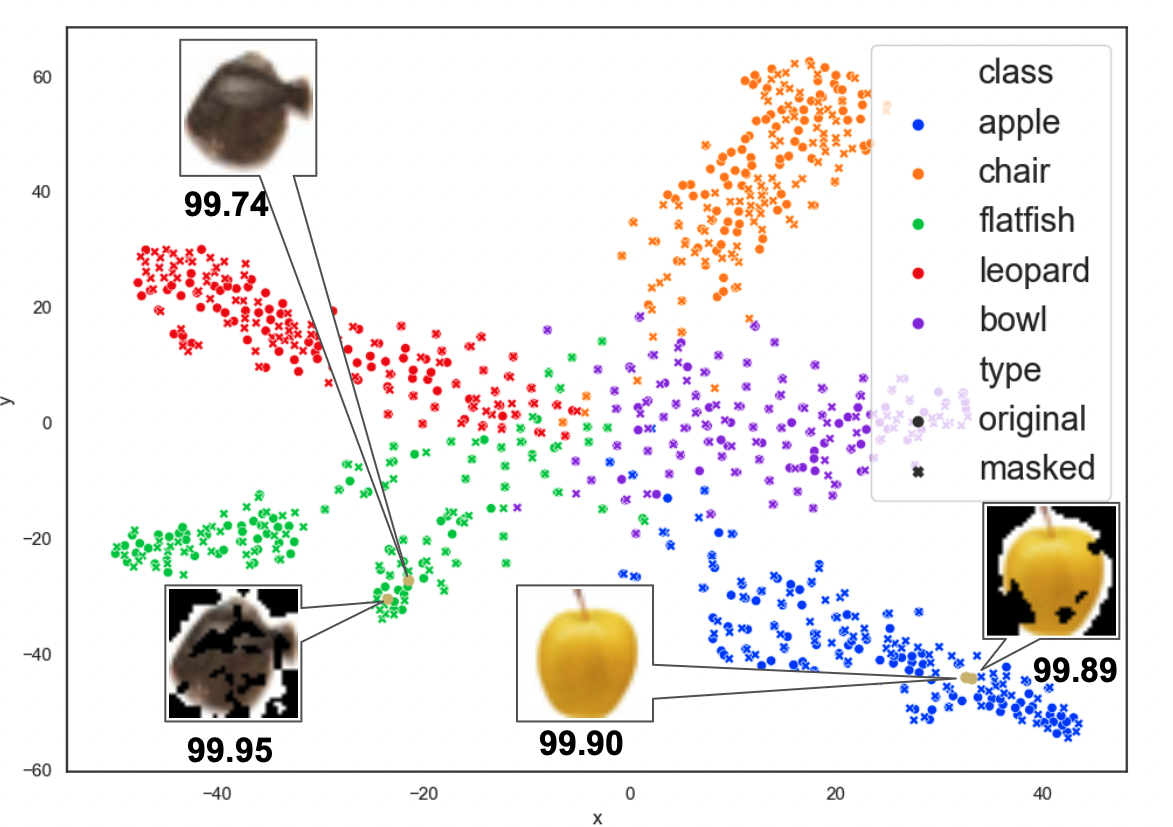}

    \vspace{-0.2cm}
    \caption{Example t-SNE embeddings for SupCon -- \emph{baseline} (left) and ExConB -- \emph{ours} (right) on the CIFAR-100 dataset.  There are five different classes on the graph, where each color represents a different class label. The cross (X) points represent the embeddings for masked versions of the input instances based on explanations, while the dots represent the embeddings for original input instances. The number below an input image indicates the probability (as a percentage) of the predicted class from a softmax layer. We make three observations when comparing ExConB to SupCon: (1) The ExConB embeddings for instances associated with different classes are farther apart from each other; (2) For instances within the same class, ExConB embeddings between original images and their masked versions are much closer; (3) ExConB prediction probabilities of the correct classes are maintained for masked images in contrast to SupCon. This illustrates the capability of ExConB to better capture class-discriminative semantic image content in its embeddings.}
    \label{fig:baseline_embedding_cifar100}
\end{figure*}

Contrastive learning has led to significant improvement in the field of unsupervised representation learning by encouraging nearby embeddings for similar data instances and distant embeddings for dissimilar instances \cite{tian2020contrastive,he2020momentum,chen2020simple,chen2020big}. Recently, \emph{supervised contrastive learning}~\cite{khosla2020supervised} has incorporated label information into contrastive representation learning. Within each training data batch, the supervised contrastive loss encourages images from the same class to have similar embeddings while pushing apart image embeddings from distinct classes. Empirically, supervised contrastive learning outperforms supervised cross-entropy learning in terms of classification accuracy. However, supervised contrastive learning relies heavily on random data augmentations (e.g., random cropping) to produce similar views of the same image \cite{khosla2020supervised}, which may cause the loss of semantic information in the original image.  In turn, such a loss of semantic image content may then lead to poor contrastive embeddings and degraded downstream task performance.

We argue that if an augmentation is performed such that it takes into account parts of the input whose semantics match the information provided by the label, it could significantly improve the performance of the model for the task at hand as well as increase its robustness to non-semantic shifts (e.g., image noise) in the input distribution. This could be explicitly done by leveraging local explanation methods \cite{simonyan2013deep,smilkov2017smoothgrad,kim2018interpretability,selvaraju2017grad,chattopadhay2018grad,ramaswamy2020ablation,lundberg2017unified,sundararajan2017axiomatic}, where the model’s output for an individual data instance is explained based on feature (e.g., pixel-level) importance.

To this end, we propose an explanation-driven supervised contrastive learning framework (ExCon) using two augmented views, where one augmentation is generated taking into account parts of the input that explain the model's decision for a given data example. This can be achieved using feature attribution methods such as GradCAM \cite{selvaraju2017grad}. Qualitatively, as shown in Figure \ref{fig:baseline_embedding_cifar100}, we have observed closer embeddings among similar examples (i.e., both between original images and their masked versions and between images of the same class) and more distant embeddings among different classes. Moreover, the performance of ExCon(B) also suggests it can better capture class-discriminative information compared to the SupCon baseline given it's substantially improved confidence scores for the correct prediction on masked images.
Quantitatively, we show that our method outperforms the supervised contrastive learning baseline introduced in \cite{khosla2020supervised} in terms of classification accuracy, explanation quality, and adversarial robustness as well as calibration of probabilistic predictions of the model in the context of distributional shift of encoder representations. 

In summary, we outline the following key contributions of our proposed ExCon(B) methodology: 
\begin{enumerate}
    \item We propose an explanation-driven supervised contrastive learning framework ExCon(B) that aims to preserve the original image semantic content when producing image augmentations for positive examples. Such explanation-driven augmentation guides the encoder to learn better semantic embedding representations for downstream tasks. We also propose an adapted loss formulation (ExConB) for negative examples to encourage the encoder to avoid semantically irrelevant content in the input image.

    \item We verify that ExCon(B) provide better quality representations as indicated by consistently better classification accuracy compared to the supervised contrastive learning baseline. We also verify that ExCon(B) provides more faithful explanations through an analysis of \emph{drop} and \emph{increase} scores. 
    \item Finally, we observe stronger adversarial robustness as well as better probabilistic calibration under distributional shift for ExCon(B) compared to the baselines.
\end{enumerate}

\section{Related Work}\label{relatedwork}

\noindent \textbf{Contrastive Representation Learning} has seen a plethora of work that has led to state-of-the-art results in self-supervised learning \cite{oord2018representation,hjelm2018learning,tian2020contrastive,arora2019theoretical,chen2020simple}. In the absence of labels, self-supervised contrastive learning relies on selecting positive pairs for each original input example. The formation of positive pairs is performed through data augmentation based on the original image \cite{chen2020simple,henaff2020data,hjelm2018learning,tian2020contrastive}. Negative examples are drawn uniformly randomly from the dataset. It is assumed that such sampling would result in an insignificant number of false negatives. An encoder network is then pretrained to discriminate between these positive and negative pairs. This pretraining allows the encoder to learn the structure of the data by encoding positive examples closer to each other in the embedding space while distancing the negative ones and pushing them apart. Once pretrained, the encoder could be used later for downstream tasks. It is clear that in the context of contrastive learning, the formation of positive and negative pairs (by random augmentations and uniform sampling respectively), does not take into account the semantics of the input that are relevant to the downstream tasks. \\

\noindent \textbf{Supervised Contrastive Learning} \cite{khosla2020supervised} leverages label information to bring contrastive learning into the supervised setup where in the embedding space, instances of the same class are clustered together while instances of distinct classes are spread apart. \cite{taghanaki2021robust} trains a transformation network through a triplet loss \cite{schroff2015facenet,koch2015siamese}, which aims at reducing irrelevant information in the input data. It is important to mention here that the framework proposed in \cite{taghanaki2021robust} cannot be compared with supervised contrastive learning frameworks such as ours (ExCon(B)) and the one proposed in \cite{khosla2020supervised} as it doesn't rely on an encoder pretraining approach. 

\noindent\textbf{Feature Attribution Methods for Local Explanations} To make sure that the explanation-driven augmentations relate to the task-relevant semantics of the input, one must guarantee that the provided local explanations are of good quality. A local explanation describes the model's behavior in the neighborhood of a given input example. An important number of works on local explanation rely on post-hoc methods such as LIME \cite{ribeiro2016should} and SHAP \cite{lundberg2017unified} where the goal is to measure contributions of the input features to the model's output.
The quality of a local explanation can be measured by how much it is aligned (faithful) to the model's prediction. The faithfulness aspect of an explanation reflects how accurate is an explanation in its estimation of the features' contributions to the model's decision process. 
In the context of convolutional neural networks (CNNs), gradient-based saliency map methods are commonly adopted to produce saliency-based explanations. Such post-hoc local explanations highlight the input features which contribute the most to the model's prediction for a given input instance \cite{simonyan2013deep,smilkov2017smoothgrad,sundararajan2017axiomatic,selvaraju2017grad}. Under the assumption of linearity, which states that certain regions of the input contribute more than others to the decision process of the model and that the contributions of different parts of the input are independent from each other, saliency-based explanations can be considered as faithful to the model's behavior \cite{jacovi2020towards}.

\noindent \textbf{Grad-CAM} Our proposed ExCon framework is explainer-agnostic. We can then adopt any local explanation method in order to perform data augmentation. Given the faithful aspect of saliency maps under the linearity assumption, it is then convenient to opt for a gradient-based saliency map explanation method. In our case, we choose Grad-CAM \cite{selvaraju2017grad} as our explainer. Grad-CAM provides a saliency-map visualization highlighting the most contributing regions to the model's output by examining the gradient flowing from the output to the final convolution layer. The use of Grad-CAM is convenient in the context of ExCon both because it is class-specific and cheap to run as it does not involve expensive sampling.

\noindent \textbf{Explanation-guided Learning}  \cite{selvaraju2021casting} utilizes pre-computed saliency map annotations to guide the attention of the representer in a self-supervised representation learning setting. However, such saliency map annotations require an additional separate step of neural network training, which is expensive and limits the application of this method. \cite{li2021edda} applies an explanation-driven augmentation strategy in a cross-entropy setting, which improves the explanation faithfulness but harms the classification performance of the model. \cite{kim2020puzzle} utilizes saliency information to generate guided-mixup augmentations. However, this method involves a separate optimization procedure to calculate the optimal mixup parameters, which could be computationally expensive in practice. Relying on concept-based explanations \cite{kim2018interpretability},  \cite{wickramanayake2021explanation} incorporates under-represented concepts that lead to misclassification and enlarge the dataset. However, the algorithm involves accumulating additional data instances along the entire training process, which can cause an increasing amount of computational resources along the training iterations.
To the best of our knowledge, our proposed framework is the first to explore explanation-driven augmentations in the supervised contrastive learning setup without using additional object localization information.

\section{Methodology}\label{tr_pipeline}

\begin{figure*}[t!]
    \centering
    \includegraphics[width=1\textwidth]{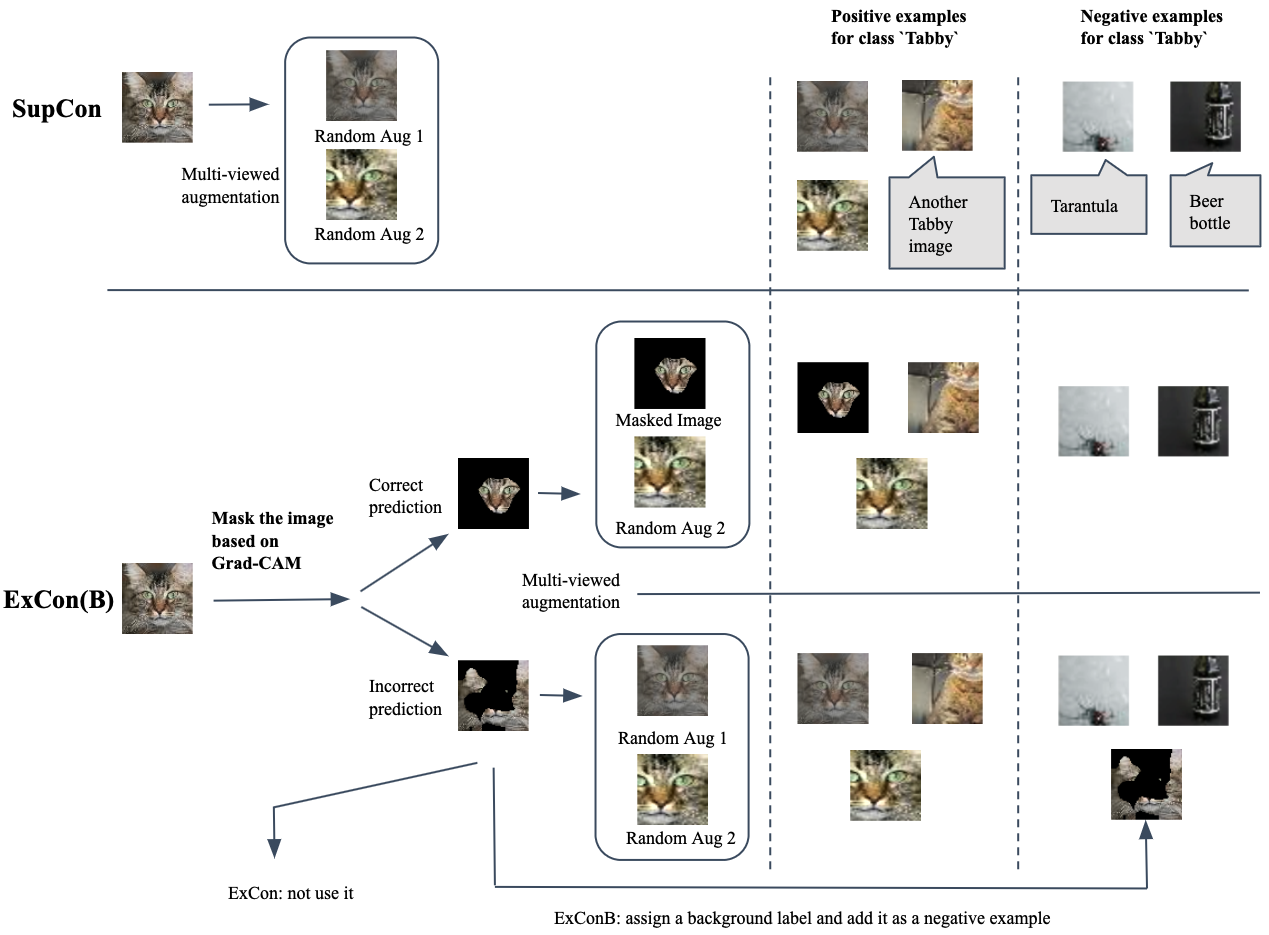}
    \caption{Methodology visualization for SupCon (baseline) and ExCon(B). The left-hand side is the formation of the multi-viewed augmentation for input images for SupCon and ExCon(B) respectively. The right-hand side is the visualization of the set of positive examples and negative examples accordingly for each scenario.}
    \label{fig:methodology}
\end{figure*}

In this section, we first revisit the supervised contrastive learning baseline and then introduce the proposed ExCon and ExConB methods.

\subsection{Supervised Contrastive Learning Preliminaries}

The major pipeline of supervised contrastive learning \cite{khosla2020supervised} has the following components:
\begin{itemize}
    \item \textbf{Augmentation} \cite{khosla2020supervised} relies on random operations (random cropping, random horizontal flipping, random color jittering, and random grayscale) to form multi-viewed augmentations. As shown in the augmentation pipeline in Figure \ref{fig:methodology}, for an input instance $\bm{x}_i$ in the dataset $\mathbf{I} = \{\bm{x}_{1}, \bm{x}_{2},  ..., \bm{x}_{n-1}, \bm{x}_{n}\}$ of size $n$, two randomly augmented instances (views) $\bm{\tilde{x}}_{2i-1}$ and $\bm{\tilde{x}}_{2i}$ are obtained. This procedure is repeated for each input instance in the dataset $\mathbf{I}$ in order to generate a multiviewed augmentation set $\mathbf{I}_{SupCon}$ of size $2n$ where $\mathbf{I}_{SupCon} = \{\bm{\tilde{x}}_{1}, \bm{\tilde{x}}_{2},  ..., \bm{\tilde{x}}_{2n-1}, \bm{\tilde{x}}_{2n}\}$.
    \item \textbf{Encoder} Any augmented input image $\bm{x}$ is fed into the encoder network to obtain the representation $\bm{r} = \mathbf{Enc}(\bm{x})$.
    \item \textbf{Projection Head} The representation $\bm{r}$ is then fed into the projection network (typically a ReLU-based multi-layer perceptron) and an L2 normalization to get the normalized projected embedding $\bm{z} = normalization(\mathbf{Proj}(\bm{r}))$.
    \item \textbf{Classification Prediction} The encoder $\mathbf{Enc}(\cdot)$ and the linear classifier $\mathbf{Clf}(\cdot)$ is combined to make the prediction during test time. For an input $\bm{x}$, the softmax activation scores are $\bm{h} = softmax[\mathbf{Clf}(\mathbf{Enc(\bm{x}}))]\in \mathbb{R}^C$ where $C$ is the total number of classes. The prediction is $\hat{\bm{y}} = \argmax_c \bm{h}$.
\end{itemize}
 \textbf{Supervised Contrastive Loss} In supervised contrastive learning, every input example in the data batch will be used as an anchor. For each particular anchor $\bm{\tilde{x}}_i\in \mathbf{I}_{SupCon}$, as shown in Figure \ref{fig:methodology}, the set of instances with the same label form positive examples, denoted by $\mathbf{P}(i)$, and those that have different labels form negative examples. Formally, the supervised contrastive learning loss \cite{khosla2020supervised} is expressed as the following, where $\tau$ is the softmax temperature:
\begin{equation}\label{supcon_loss}
\scalebox{0.8}{
    $\mathcal{L}^{SupCon} = \displaystyle{\sum_{i\in \mathbf{I}_{SupCon}}}\frac{-1}{|\mathbf{P}(i)|} \sum_{p\in \mathbf{P}(i)} \log \left\{ \frac{\displaystyle\mathrm{exp}(\bm{z}_i\cdot \bm{z}_p / \tau)}{\displaystyle \sum_{a\in \left[\mathbf{I}_{SupCon}\textbackslash \{i\}\right]}\mathrm{exp}(\bm{z}_i\cdot \bm{z}_a / \tau)}\right\}$},
\end{equation}

\subsection{ExCon and ExConB}\label{methodology}

We build upon the supervised contrastive learning framework presented in \cite{khosla2020supervised}. We perform explanation-driven data augmentation to encourage the model to consider the  task-relevant features in its decision-making process. We also propose a new formulation of the loss function which allows the model to take into account negative examples which are not part of the set of anchors. \\

\noindent\textbf{ExCon} As visualized in Figure \ref{fig:methodology}, for each input image, an explanation-driven augmented version is acquired by masking the pixels that have a Grad-CAM importance score less than the threshold 0.5. If the masked image is of high quality, i.e., if the masked image yields a correct prediction, then together with a randomly augmented version, they are adopted as the multi-viewed pair for the original image. If the masked image is of bad quality, i.e., if it yields an incorrect prediction, then only two randomly augmented versions will be used to form the multi-viewed pair. The set of augmentations in the case of ExCon is denoted as $\mathbf{I}_{ExCon}$ where $|\mathbf{I}_{ExCon}| = |\mathbf{I}_{SupCon}| = 2*n$. Since there are no extra images introduced in ExCon, it simply follows the supervised contrastive loss.\\

\noindent\textbf{ExConB Loss}\label{exconb} As visualized in Figure \ref{fig:methodology}, in the case where the masked image yields a correct prediction, the pipeline is identical to ExCon. In the case where the masked image yields an incorrect prediction, we want the model to become \emph{implicitly} aware of the internal mechanisms that mislead it to erroneous decisions. Hence we assign the masked image a background label and add two of it (in order to form a pair) to the batch. Given that background images originated from different images could possibly have different semantics, we don't use the background images as anchors such that they are neither matched nor contrasted with each other. As a result, the only role for the background images is negative examples to the anchors in the data batch. We denote the set of background masked images in the batch using $\mathbf{B}$. Therefore, $\mathbf{I}_{ExConB} = \mathbf{I}_{ExCon}\cup \mathbf{B}$

However, this functionality cannot be built directly upon the supervised contrastive loss, since supervised contrastive loss will treat every example in the batch as an anchor. As a solution to mitigate this limitation, we use $\mathbf{I}_{ExCon}$ as the anchor set and $\mathbf{I}_{ExCon}\cup \mathbf{B}$ as the normalization set. Formally, we introduce the ExConB loss as follows:

\begin{equation}\scalebox{0.75}{
    $\mathcal{L}^{ExConB} = \displaystyle{\sum_{i\in \mathbf{I}_{ExCon}}}\frac{-1}{|P(i)|} \sum_{p\in P(i)} \log \left\{ \frac{\displaystyle\mathrm{exp}(\bm{z}_i\cdot \bm{z}_p / \tau)}{\displaystyle \sum_{a\in \left[(\mathbf{I}_{ExCon}\cup \mathbf{B})\textbackslash \{i\}\right]}\mathrm{exp}(\bm{z}_i\cdot \bm{z}_a / \tau)}\right\}$.
}\end{equation}

\noindent\textbf{Training Pipeline for ExCon(B)} Note that we want to take advantage of the explanation-driven augmentations in such a way that it reflects the changing behavior of the entire model during the training process. As the model is composed of the encoder and the classifier, providing the former with the non-stationary explanations of the classifier's outputs during training allows it to keep track of the changing behavior of the whole model and how it affects its decision-making process through the training iterations. For this reason, we iterate between the encoder $\mathbf{Enc}(\cdot)$ and the classifier $\mathbf{Clf}(\cdot)$ at each training epoch.

\section{Experiments}\label{experiments}

We design the experimental work to answer the following research questions:
\begin{enumerate}
    \item \textbf{Research Question 1}: Does our method ExCon(B) improve the encoder representations? 
    Since representation quality is measured by classification scores through combining the encoder with a linear classifier \cite{arora2019theoretical}, we are interested in exploring the classification accuracy of ExCon(B) and the supervised contrastive learning baseline.
    \item \textbf{Research Question 2}: Does ExCon(B) improve the explanation quality of the trained models given that they are trained explicitly based on the feature attribution explanation method?
    \item \textbf{Research Question 3}: Does our method ExCon(B) improve the adversarial robustness of the model?
    \item \textbf{Research Question 4}: Compared to two-stage training, how does the distributional shift of encoder representations induced by iterative training affect the probabilistic calibration of a linear classifier?
\end{enumerate}

\begin{figure*}[t!]
    \centering

        \includegraphics[width=0.51\textwidth]{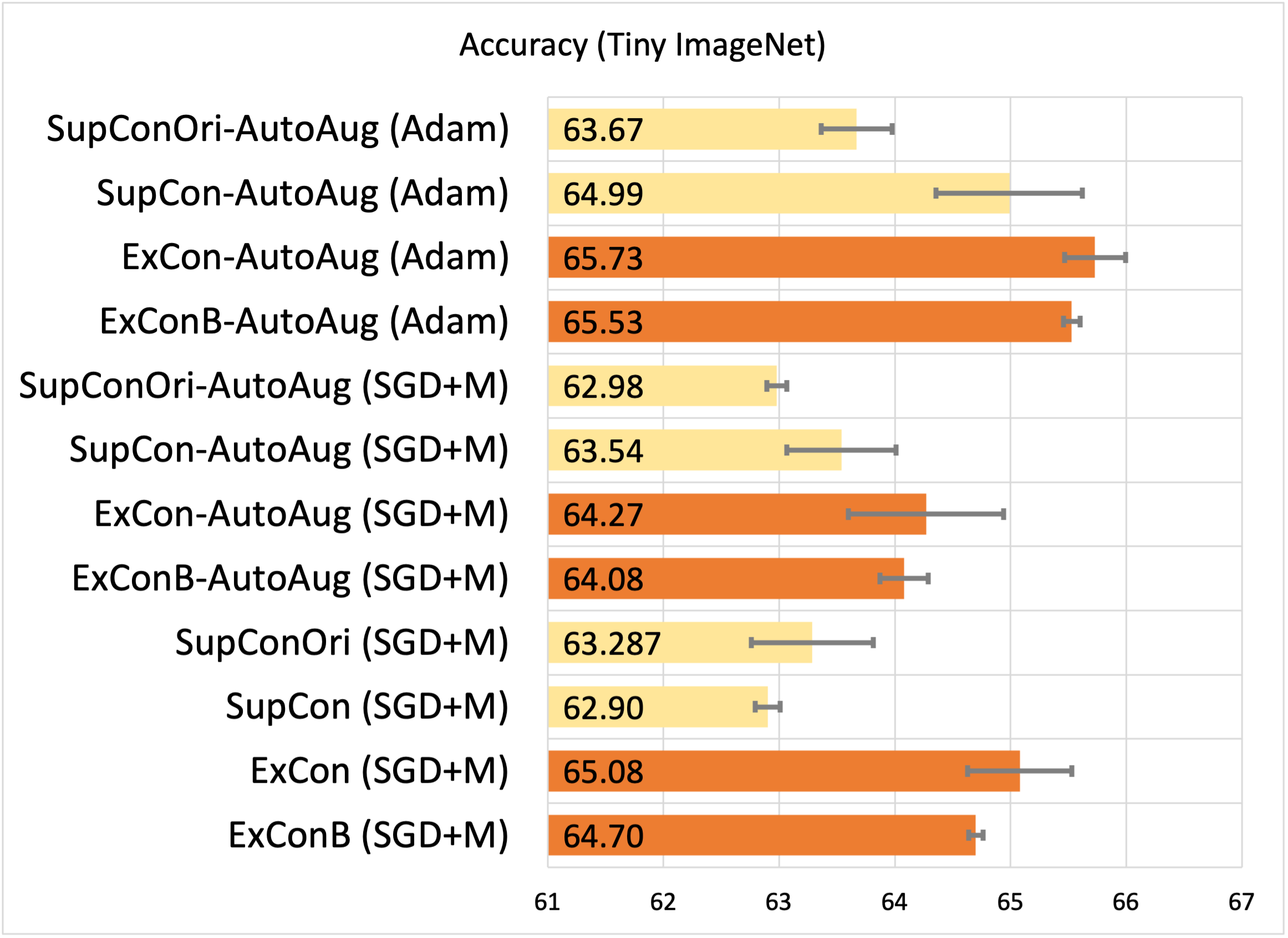}
        \includegraphics[width=0.46\textwidth]{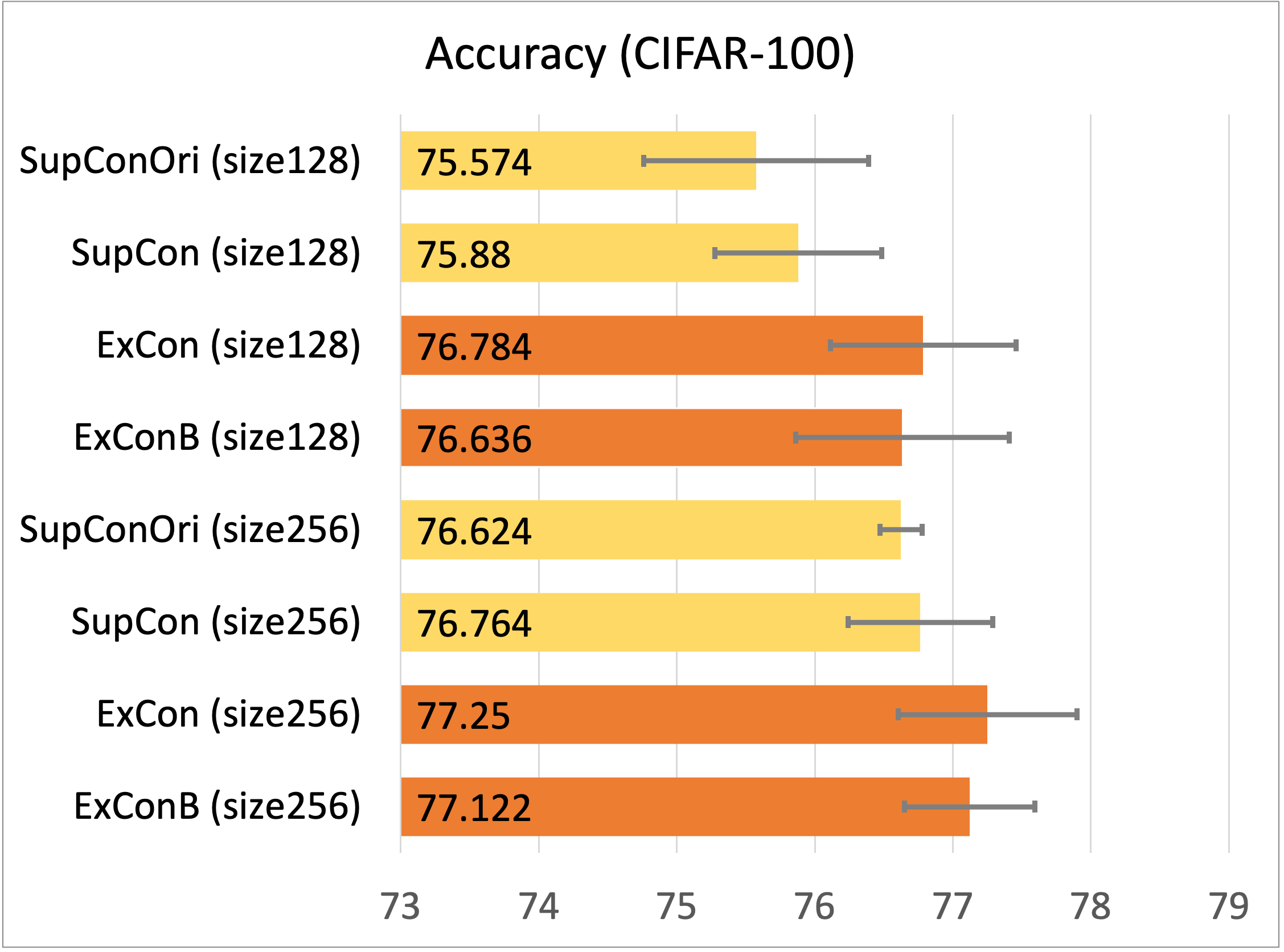}
        
    \caption{Classification accuracies: 1. Tiny ImageNet 350 epoch models on the left 2. CIFAR-100 (batch sizes 128 and 256) models on the right. ``SGD+M" stands for SGD with momentum optimizer.}
    \label{fig:classification_main_text}
\end{figure*}

\subsection{Experimental Settings}
Here we outline the methods as well as the datasets used for the experiments. We provide full training details into the Appendix.\\

\noindent\textbf{Datasets} We used the Tiny ImageNet \cite{chrabaszcz2017downsampled} as well as the CIFAR-100 \cite{krizhevsky2009learning} datasets to verify the efficiency of our methods. For the Tiny ImageNet dataset, we adopted a batch size of 128 and trained for 350 epochs. For the CIFAR-100 dataset, we adopted batch sizes of 128 and 256 and trained for 200 epochs. We could not adopt larger batch sizes due to resource limitations. For both datasets, we have tried SGD with the momentum optimizer. For Tiny ImageNet, we have addtionally tried the Adam optimizer \cite{kingma2014adam}. Due to space limitations, we only provide the portion of results relevant to each research question. Complete results are included in the Appendix.\\

\noindent\textbf{Methods for Comparison}
\begin{itemize}
    \item \textbf{SupConOri}: The supervised contrastive learning method displayed in \cite{khosla2020supervised}, where the encoder is fully trained first and then the classifier is trained based on the fixed encoder.
    \item \textbf{SupCon}: Same as SupConOri except that both the encoder and classifier are trained accordingly under each training iteration. This is for fair comparison with ExCon and ExConB where the same iterative training procedures are adopted.
    \item \textbf{ExCon}: As introduced in \ref{methodology}.
    \item \textbf{ExConB}: As introduced in \ref{methodology}.
    \item \textbf{SupCon with AutoAugment}: Both of the augmented views of the original image come from the AutoAugmemt \cite{cubuk2018autoaugment} policy. Here instead of training a reinforcement learning framework, we used the pre-trained policy from ImageNet. Note that we only experimented with the AutoAugment strategy on the Tiny ImageNet dataset 
    since the CIFAR-100 dataset benefits more from simple augmentation as noted in the SupCon repo~\cite{khosla2020supervised}.
    \item \textbf{SupConOri with AutoAugment} The non-iterative training strategy for the supervised contrastive learning baseline with the AutoAugment strategy.
    \item \textbf{ExCon(B) with AutoAugment}: For each original image, if the masked image yields a correct prediction, the augmented pair composes the masked image and another image coming from AutoAugment. Otherwise, the augmented pair composes two AutoAugment images.
\end{itemize}

\subsection{Research Question 1: Classification Accuracy}\label{RQ1}
As discussed previously, the classification accuracy is a major indicator of encoder representation quality \cite{arora2019theoretical}.

From the results reported in Figure \ref{fig:classification_main_text}, we can observe that the ExCon (B) methods generalize better than SupCon(Ori) on the CIFAR-100 classification task. The gaps in terms of top-1 classification accuracy are even more significant for the Tiny ImageNet classification task. This suggests a fact that when the task in question involves higher resolution input (more input features), more examples, and a larger number of classes, the job of capturing task-relevant semantics and separating it from the spurious features becomes easier for the model. The capability of the ExCon(B) methods to (a) take into account task-relevant features in order to embed representations of the same class (as well as their explanations) closer to each other in the embedding space and (b) push apart representations of distinct classes helps the model perform the classification task at hand. This is illustrated in the t-SNE plots \cite{van2008visualizing} shown in Figure \ref{fig:baseline_embedding_cifar100}.

\begin{figure}[h]
    \centering
        \includegraphics[width=0.45\textwidth]{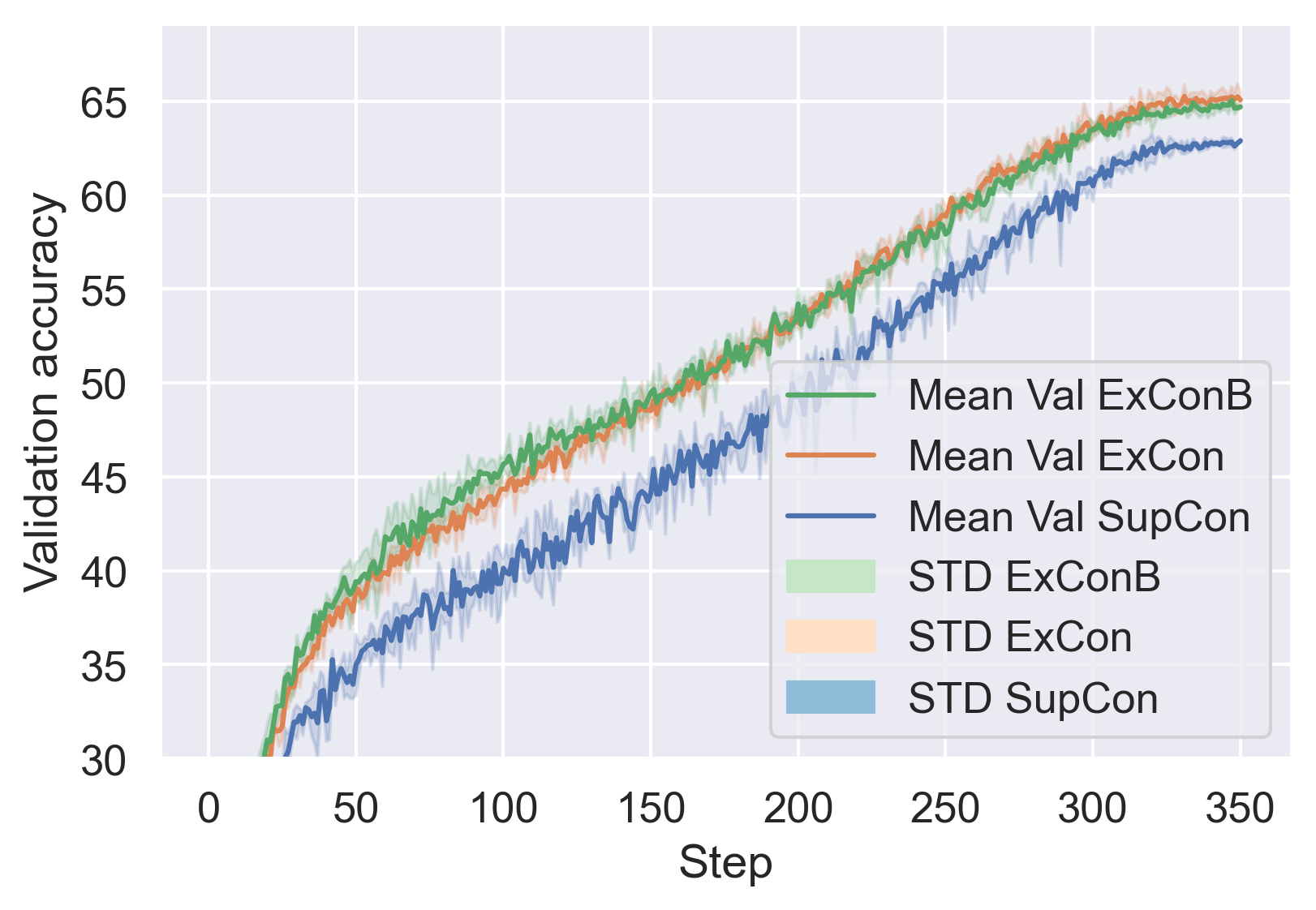}

    \caption{Average validation accuracies along the training epochs on the Tiny ImageNet dataset. We can see that the training related to our proposed methods is more stable than the baseline, which can be verified through the standard deviation over 3 random seeds.}
    \label{fig:main_convergence}
\end{figure}

It is important to mention that smoother validation accuracy curves with reduced variance are obtained for the ExCon(B) methods compared to the SupCon baseline under iterative training. This can be observed in Figure \ref{fig:main_convergence}. As the  variance reduction of the classification accuracy is consistent throughout the training epochs, this suggests that the model is more stable for the ExCon(B) methods when the iterative training procedure is adopted compared to the SupCon baseline. The effect of introducing explanation-driven augmentation into the training pipeline and the iterative training procedure on the calibration of the model is investigated more in depth in Section \ref{RQ4}.

\subsection{Research Question 2: Explanation Quality}\label{RQ2}

\begin{figure*}[t!]
    \centering
    \includegraphics[width=0.49\textwidth]{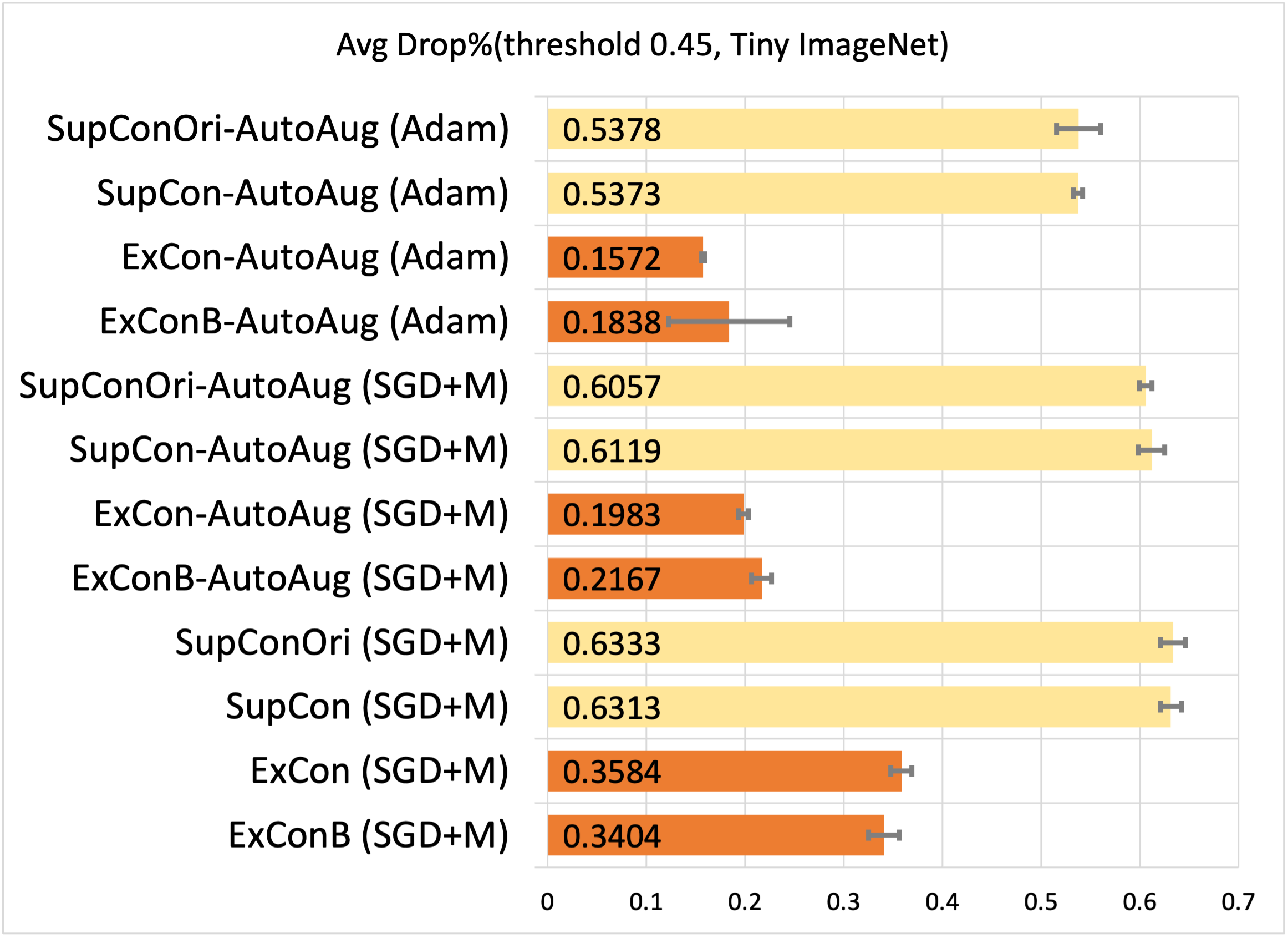}
    \includegraphics[width=0.49\textwidth]{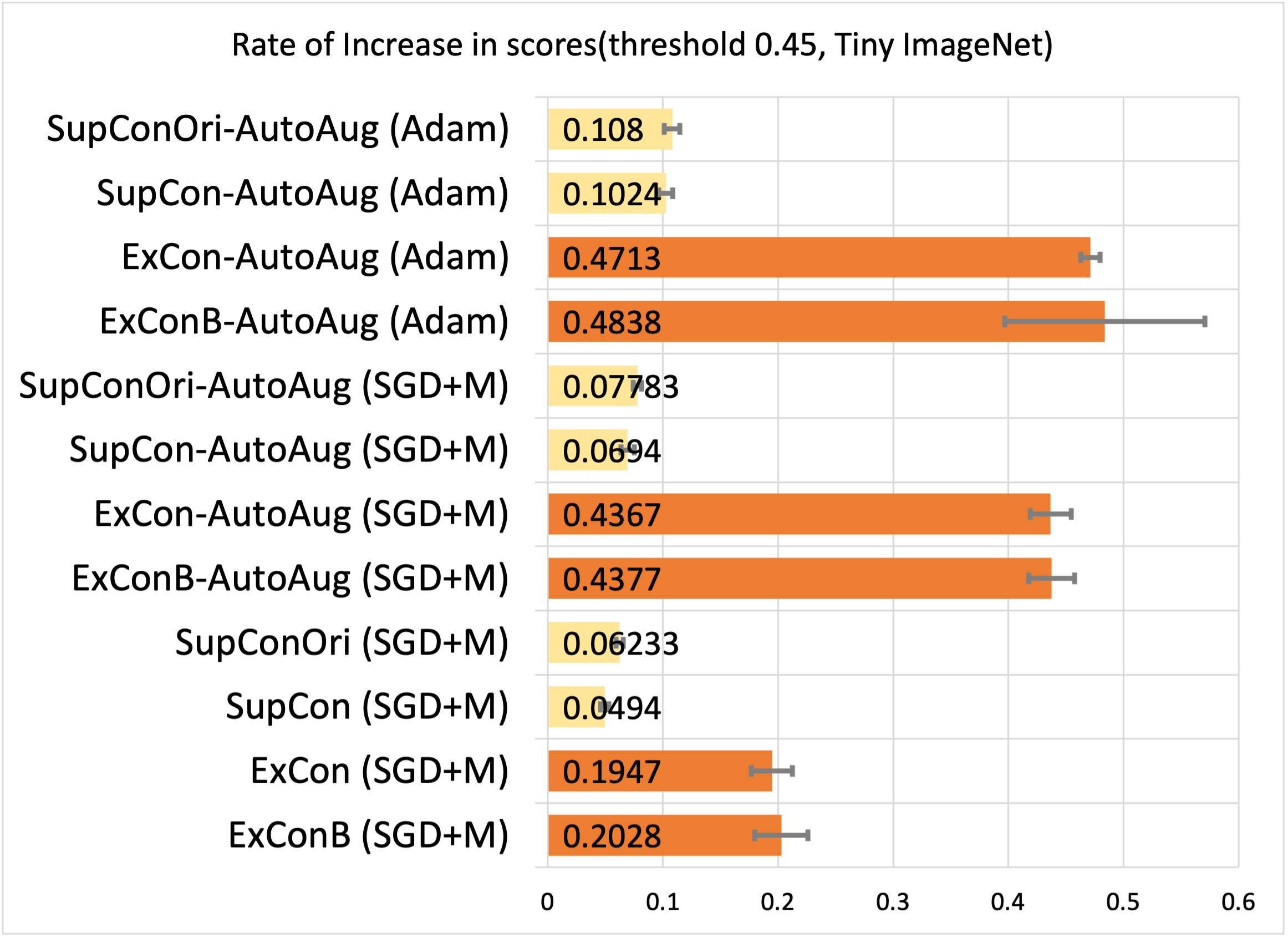}
    \caption{Average Drop \% (lower is better) and Rate of Increase in Scores (higher is better) of the models trained on Tiny ImageNet for 350 epochs.}
    \label{fig:main_text_drop_increase}
\end{figure*}

\noindent \textbf{Drop/Increase Scores} An explanation of good quality highlights the parts of the input that contribute the most to the neural network's decision. This property is called explanation faithfulness, as introduced in \cite{jacovi2020towards}. When fed in as the input to the model, the masked version that reserves only the important regions of the image should witness either a large increase in the confidence scores or a small decrease compared to the confidence score of the original image. Based on this intuition, the authors in \cite{chattopadhay2018grad,ramaswamy2020ablation} introduced several metrics, \textit{the average drop percentage} and \textit{the rate of increase in scores} to evaluate the explanations. The average drop percentage is defined as:
$Average\,drop\,\% = \frac{1}{T} \sum_{i=1}^{T} \frac{\mathrm{max}(0, \bm{h}_{i}^{c} - \bm{h}_{i}^{'c})}{\bm{h}_{i}^{c}} \times 100$, where $T$ is the number of test examples. $h_{i}^{c}$ denotes the softmax score when the $i^{th}$ test example is fed as input to the network. $h_{i}^{'c}$ refers to the softmax score obtained when the masked image based on the explanation of the $i^{th}$ test example is provided as input to the network. We also report the rate of increase in scores which can be defined as $Rate\,of\,increase\,in\,scores= \sum_{i=1}^{T} \frac{\mathbbm{1}_{\bm{h}_{i}^{c} < \bm{h}_{i}^{'c}}}{T} \times 100$, where $\mathbbm{1}$ is the indicator function. The rate of increase in scores is complementary to the rate of drop in scores as the two sum to one. It is therefore sufficient to report results for the rate of increase in scores. The average increase percentage is also reported and is defined as $Average\,increase\,\% = \frac{1}{T} \sum_{i=1}^{T} \frac{\mathrm{max}(0, \bm{h}_{i}^{'c} - \bm{h}_{i}^{c})}{\bm{h}_{i}^{c}} \times 100$, and due to space limitation, this is put into Appendix.\\

\noindent \textbf{Observations} The measures in question quantify the change in the softmax probability for a given class after masking unimportant regions of the input data. We report the above metrics by retaining the top-45\% of the pixels in an input image according to its corresponding explanation. Top-15\% and top-30\% results are included in the Appendix. Larger increases and smaller drops reflect better explanation quality. We can deduce from the results reported in Figure \ref{fig:main_text_drop_increase} that the ExCon(B) methods yield consistently significantly higher increase scores and lower drop measures than their SupCon(Ori) counterparts across all the models trained on CIFAR-100 and Tiny ImageNet. The ExCon(B) methods show superiority in terms of explanation quality regardless of the percentage of pixels retained in the input images.\\

\subsection{Research Question 3: Adversarial Robustness}\label{RQ3}

\begin{figure*}[t!]
    \centering
    \includegraphics[width=0.55\textwidth]{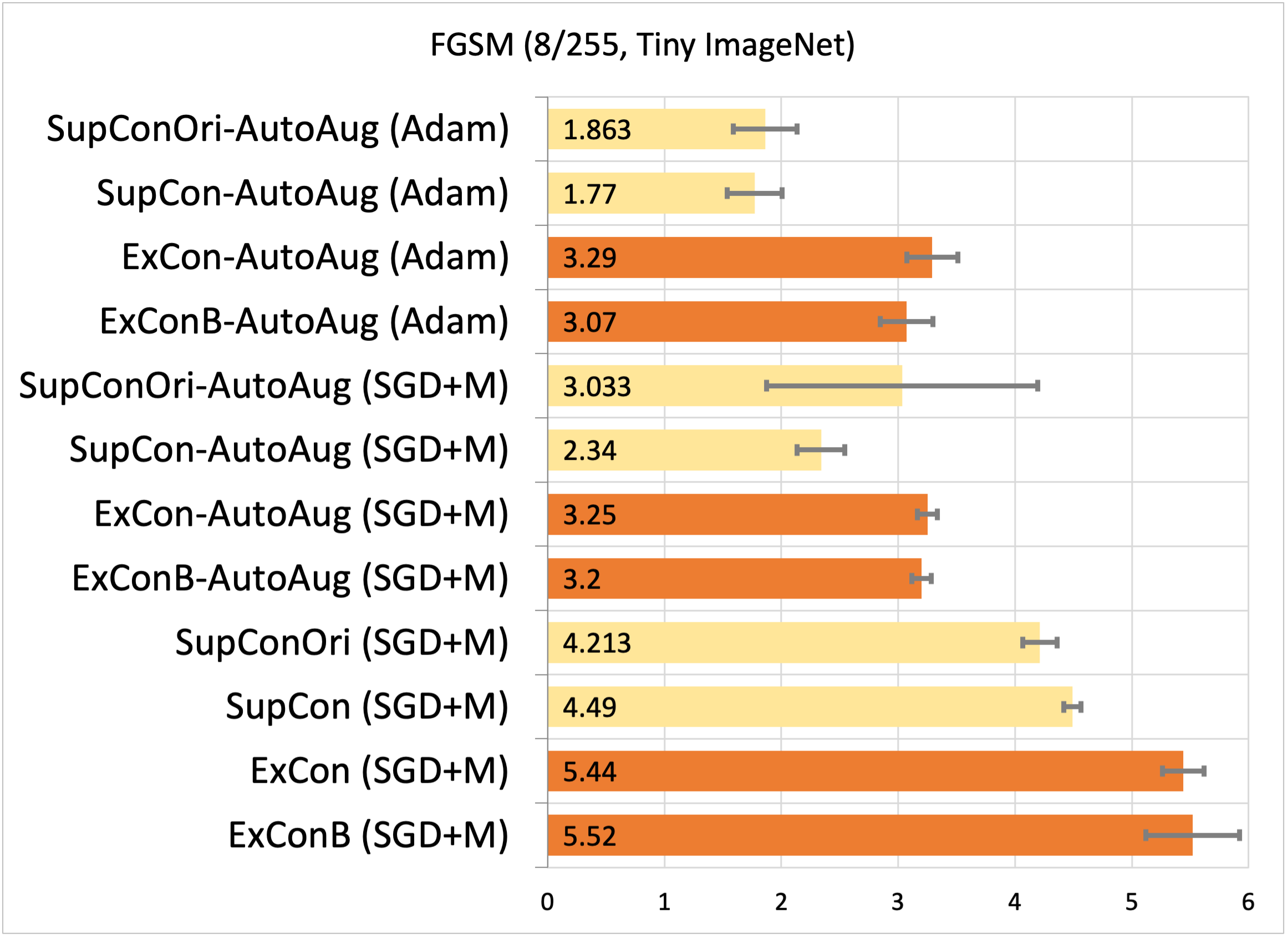}
    \includegraphics[width=0.43\textwidth]{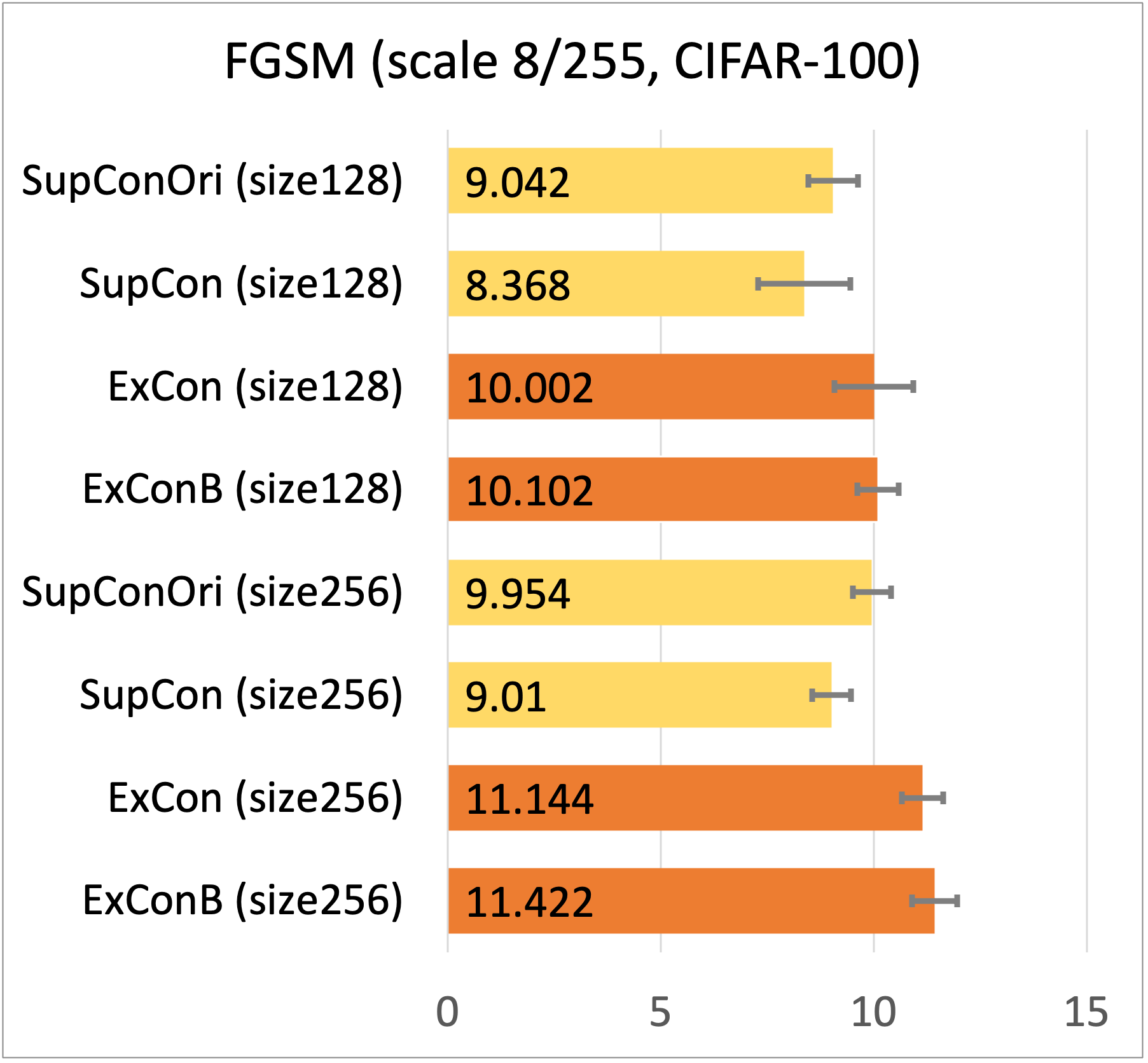}
    \caption{The top-1 accuracy under the FGSM input perturbation for the Tiny ImageNet dataset and the CIFAR-100 dataset respectively (under the scale coefficient 8/255). }
    \label{fig:main_fgsm}
\end{figure*}

\begin{figure*}[t!]
    \centering
    \includegraphics[width=0.51\textwidth]{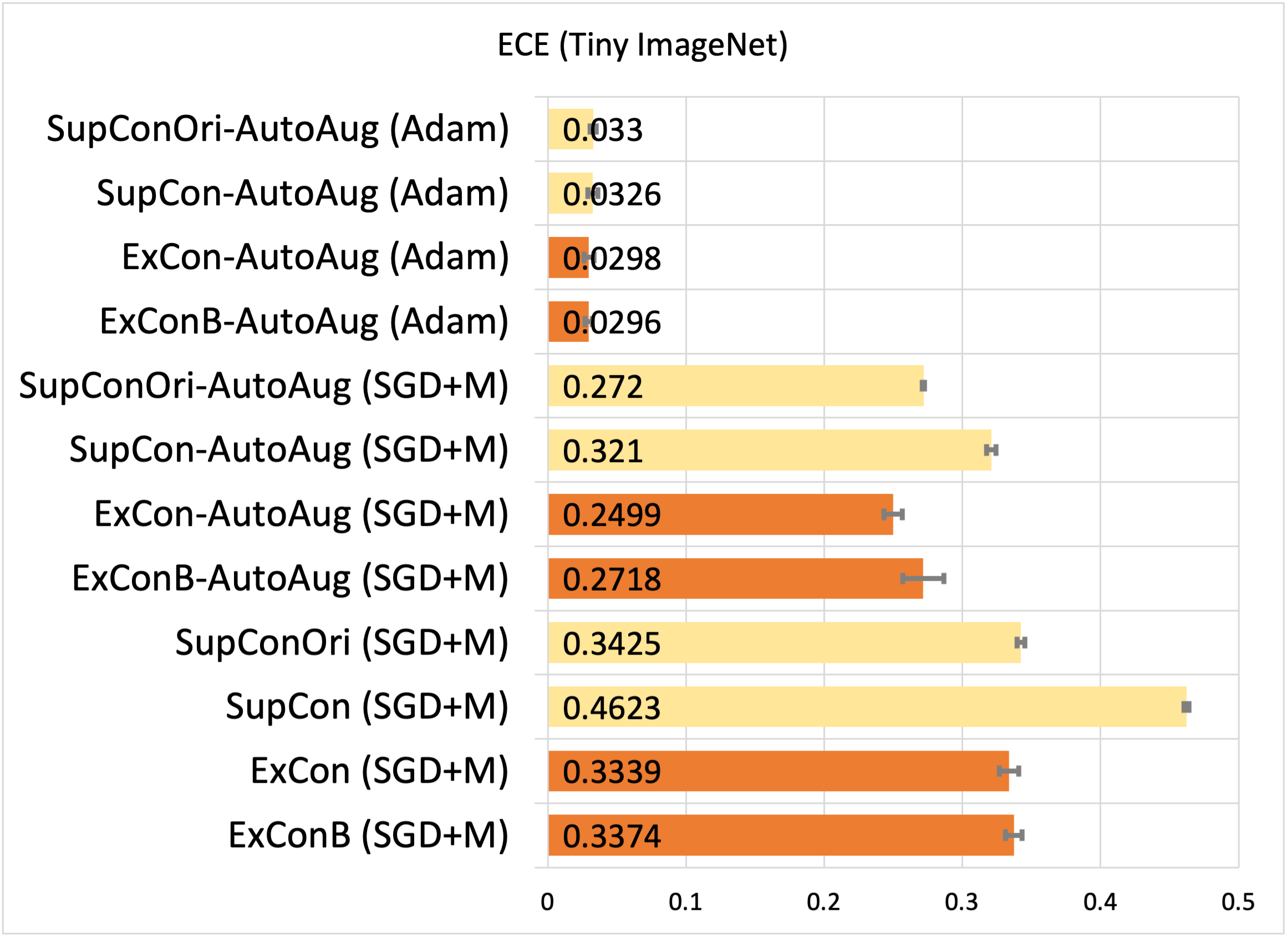}
    \includegraphics[width=0.45\textwidth]{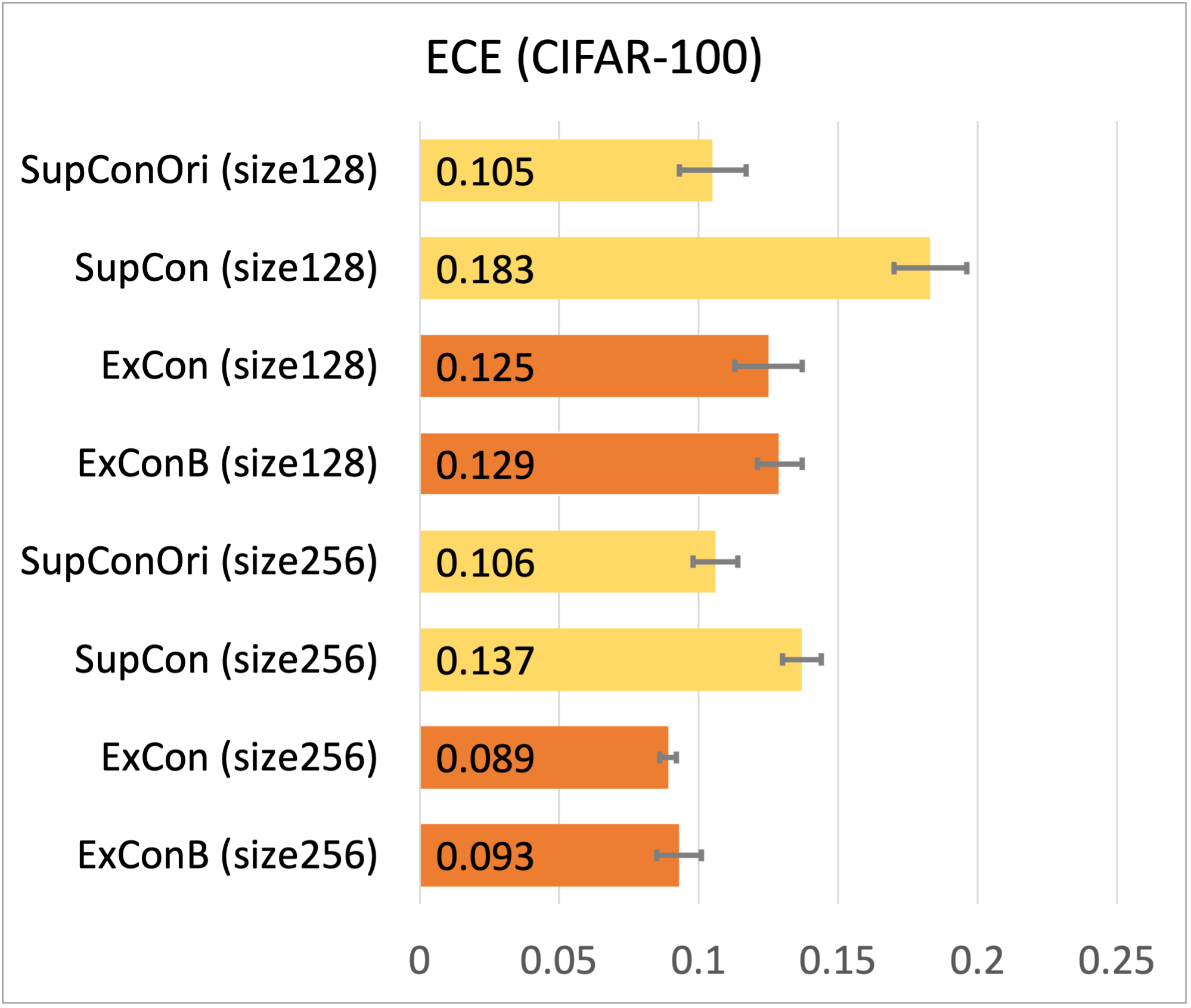}
    \caption{The expected calibration error (ECE, lower is better) for the Tiny ImageNet and the CIFAR-100 datasets.}
    \label{fig:main_ece}
\end{figure*}

\textbf{FGSM Robustness Measurement} Models that are robust to adversarial input noise maximize the distance between the input embeddings and the decision boundaries of their corresponding classes \cite{madry2017towards}. Our t-SNE plots in Figure \ref{fig:baseline_embedding_cifar100} suggest that the ExCon(B) methods are more robust to adversarial noise as the input embeddings of the same class are closely clustered together while those of different classes are pushed farther apart. This can be verified by measuring the
fast gradient sign method (FGSM) introduced in \cite{goodfellow2014explaining}. The adversarial robustness of a model is measured by adding input perturbation which is defined as the sign of the input gradient with respect to the loss function, $c\cdot sign[ \nabla_{\bm{x}} \mathcal{L}(\mathbf{\theta}, \bm{x}, \bm{y})]$, where $\mathcal{L}$ is the loss function for the network, $\mathbf{\theta}$ are the function parameters, and $c$ is the weight for controlling the scale of the perturbation. \\

\noindent \textbf{Observations} In our experiments, we follow the same weights ($c=4/255$ and $c=8/255$) as the official repo \footnote{https://github.com/snu-mllab/PuzzleMix/blob/master/main.py} of PuzzleMix \cite{kim2020puzzle}. The results for $c=8/255$ on CIFAR-100 and Tiny ImageNet are shown in Figure \ref{fig:main_fgsm}. We can see that ExCon(B) show consistently better top-1 accuracy under adversarial perturbations compared to SupCon and SupConOri. This is due to the fact that ExCon(B) is trained to focus on discriminative regions and as a consequence, it is less sensitive to the input noise.

\subsection{Research Question 4: Probabilistic Calibration under Iterative Training}\label{RQ4}

Given that SupConOri uses a fixed/trained encoder to train the linear classifier, the encoder representation distribution is therefore stationary. However, it is not the case for SupCon, ExCon and ExConB, where the encoder is updated on each iteration and the representations from the encoder shift in terms of their distribution. Hence the linear classifier has to learn to separate classes under this distributional shift.  \cite{ovadia2019can} has shown that in-distribution data can induce better calibration compared to out-of-distribution data. We aim to verify this in our scenario. \\

\noindent \textbf{Expected Calibration Error} We adopt the widely used probabilistic calibration measure -- ECE (\textit{Expected Calibration Error}) -- as introduced in \cite{guo2017calibration}. A well-calibrated model is one where, its predictive confidence score $\bm{h}$ accurately reflects its classification accuracy $\mathbb{P}(\hat{\bm{y}} = \bm{y}|\bm{h})$, where $\hat{\bm{y}}$ is the prediction and $\bm{y}$ is the ground-truth label. In order to quantify the mismatch between the two, the predictions $\hat{\bm{y}}$ are grouped into $M$ bins $B_{1}$, ..., $B_{M}$ to estimate the ECE. The accuracy and confidence score of each bin can then be calculated. The accuracy of the samples that fall into the $m^{th}$ bin $B_{m}$ is given by $acc(B_{M}) = \frac{1}{|B_{m}|} \sum_{i \in B_{m}} \mathbbm{1}_{\hat{\bm{y}_{i}} = \bm{y}_{i}}$, where $\mathbbm{1}$ is the indicator function. The average confidence for the samples that fall into the $m^{th}$ bin $B_{m}$ is given by $conf\left(B_{M}\right) = \frac{1}{|B_{m}|} \sum_{i \in B_{m}} \bm{h}_i$, where $\bm{h}_i$ is the confidence score for the $i^{th}$ sample in the $B_{m}$. An estimator of the expected calibration error can then be constructed by taking a weighted average of the absolute gaps between the bins' accuracies and their corresponding confidence scores \cite{guo2017calibration}. This is given by:
\begin{equation}
    \widehat{ECE} = \sum_{m=1}^{M} \frac{|B_{m}|}{n} \left| acc(B_{m}) - conf\left(B_{M}\right) \right|,
\end{equation} where $n$ is the total number of samples in all bins. The smaller the $\widehat{ECE}$, the more calibrated the model is.\\

\noindent\textbf{Observations} Estimates of the expected calibration error are reported for the trained models on the Tiny ImageNet and CIFAR-100 datasets in Figure \ref{fig:main_ece}. We can deduce from the reported results that when the iterative training procedure is adopted, the ExCon(B) models improve the calibration of the model as it is associated to lower errors compared to SupCon. Overall, however, calibration errors obtained for SupConOri are the lowest. This means that the encoder representation distributional shift induced by the iterative training procedure does not help calibrate the model, which verifies our assumption and is in concordance with the findings in \cite{ovadia2019can}. Consequently, the improvement in the classification accuracies and the deterioration in terms of the expected calibration error suggest continuing to train the classifier once the iterative procedure is finished. This can be investigated more in-depth in future work. Nevertheless, as shown in the Appendix, the adoption of Adam as an optimizer significantly reduces the calibration error for all the models that have been trained with an iterative procedure as the expected calibration errors become insignificantly small (within $10^{-2}$ order of magnitude).

\section{Conclusion}

We proposed a novel methodology for explanation-driven supervised contrastive learning, namely ExCon and ExConB. Qualitatively, we validated our assumption that similar examples should have similar embeddings and explanations by observing closer embeddings between images and their explanations compared to the baseline method.  Through extensive quantitative experiments, we verified that ExCon(B) outperforms supervised contrastive learning baselines in classification accuracy, explanation quality, and adversarial robustness as well as calibration of probabilistic predictions of the model in the context of distributional shift. We believe that the novel insights proposed in this paper to preserve semantic content of embeddings through the use of explanation methodologies may extend to a variety of problems in machine learning.

{\small
\bibliographystyle{ieee_fullname}
\bibliography{main}
}
\clearpage

\section*{Supplementary Materials}

\subsection*{Citation of Assets}
Our code is adapted based on the supervised contrastive learning codebase: BSD 2-Clause License \\(https://github.com/HobbitLong/SupContrast)

\subsection*{Training Details}
\textbf{Network Architecture} In all the reported experiments, we used the ResNet-50 \cite{he2016deep} as the encoder $Enc(\cdot)$ architecture for ExCon(B), as well as the SupCon baseline. Instead of using the 7$\times$7 convolutional filter like that in \cite{he2016deep} for the first convolutional layer, we adopted the 3$\times$3 convolutional filter following the PyTorch repo of SupCon \cite{khosla2020supervised} for better reserving dimensions of small images. \\

\noindent \textbf{CIFAR-100} While conducting experiments, we observed that the introduction of the explanation-driven pipeline at the start of the training procedure can hinder its convergence. For this reason, we consider the starting epoch in which we introduce the explanation-driven augmentations in the training pipeline, i.e., the ExCon(B) starting epoch, as a hyperparameter. The training data is split into train and validation subsets such that 80\% is dedicated to the train and 20\% is used for validation purposes. We used the best epoch model among the 200 epoch validation process to select the hyperparameter - starting epoch. Once the ExCon(B) starting epoch yielding the best accuracy on the validation subset is obtained, the encoder $Enc(\cdot)$ and the classifier $Clf(\cdot)$ are trained from scratch on the whole training set. All the reported results related to CIFAR-100 are obtained using models trained on 200 epochs.
As suggested in the official PyTorch repo of \cite{khosla2020supervised}, we used a learning rate of $0.5$ (with linear warmup and a cosine decay schedule \cite{loshchilov2016sgdr}), but this was for both the encoder and the classifier in our case. We adopted a softmax temperature of $0.1$. We used the stochastic gradient descent (SGD) optimizer with momentum of 0.9. \\

\noindent \textbf{Tiny ImageNet} For the Tiny ImageNet dataset, it is important to mention that no issues have been observed when introducing the explanation-driven pipeline at the start of the training procedure. This can be explained by the fact that for the Tiny ImageNet dataset, examples have higher resolution than the CIFAR-100 instances ($64 \times 64$ vs $32 \times 32$ pixels) and that both the training examples and the number of classes are significantly larger ($100$k vs $50$k training examples and $200$ vs $100$ classes) which results in an easier task for the model to separate between task-relevant semantics and irrelevant features. In fact, when explanation-driven augmentations are misclassified at the beginning of the training procedure and these are introduced as negative examples with background labels, it provides the encoder with high-resolution information which is useful to capture the spurious features that are misleading the classifier. Instead of using a learning rate of 0.5, we used a smaller learning rate of 0.2 as we are using a significantly smaller batch size compared to the ones reported in \cite{khosla2020supervised}. We also followed linear warmup as well as a cosine decay schedule \cite{loshchilov2016sgdr}.\\

\noindent Note that for both datasets, the multiviewed augmentations were only used for training the encoder.

\subsection*{Supplemental Comparative Evaluation}

Here we detail additional comparative experimental results for the quantitative analysis in the main paper as well as qualitative analysis.

\begin{figure*}
    \centering
    \includegraphics[width=0.52\textwidth]{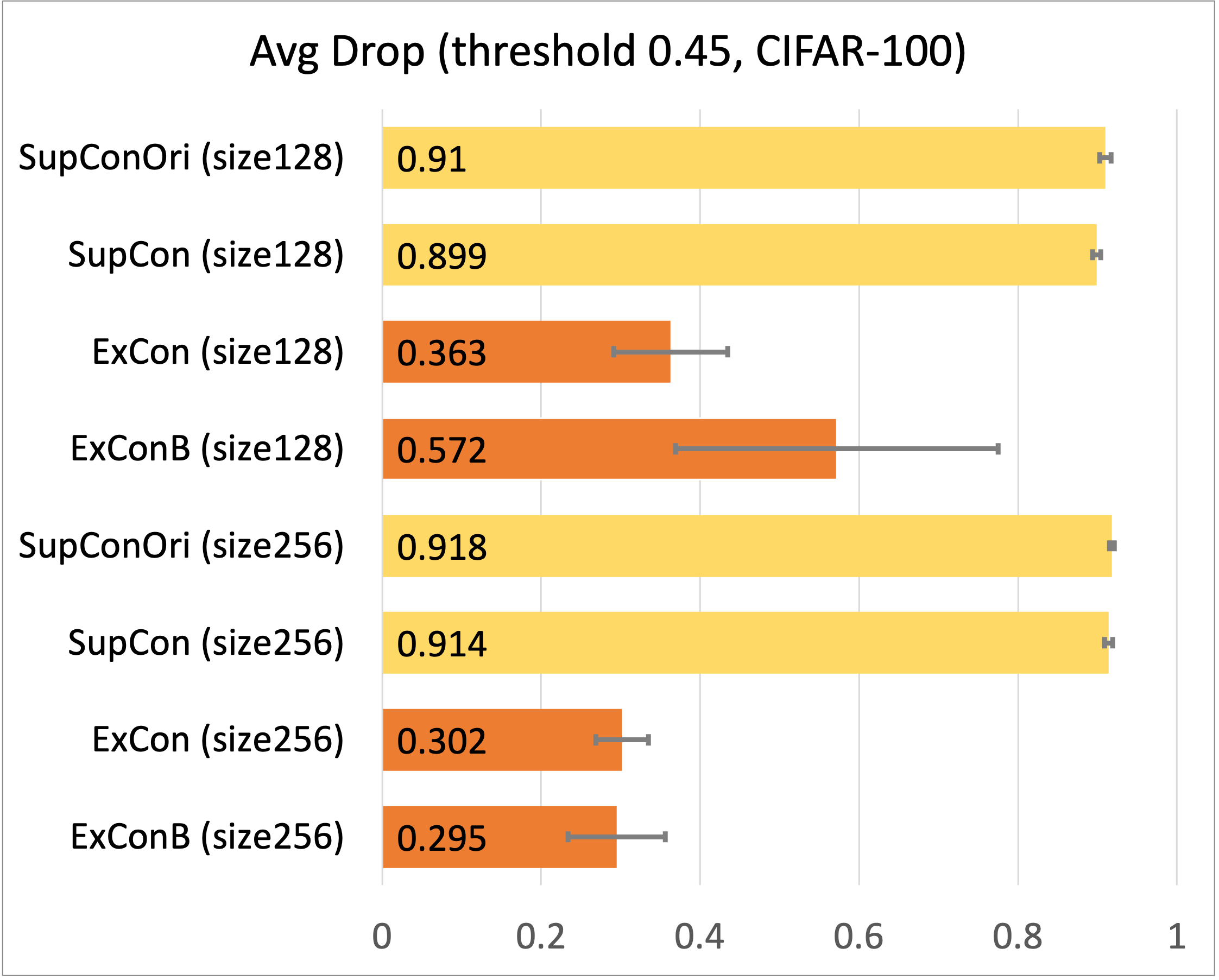}
    \includegraphics[width=0.46\textwidth]{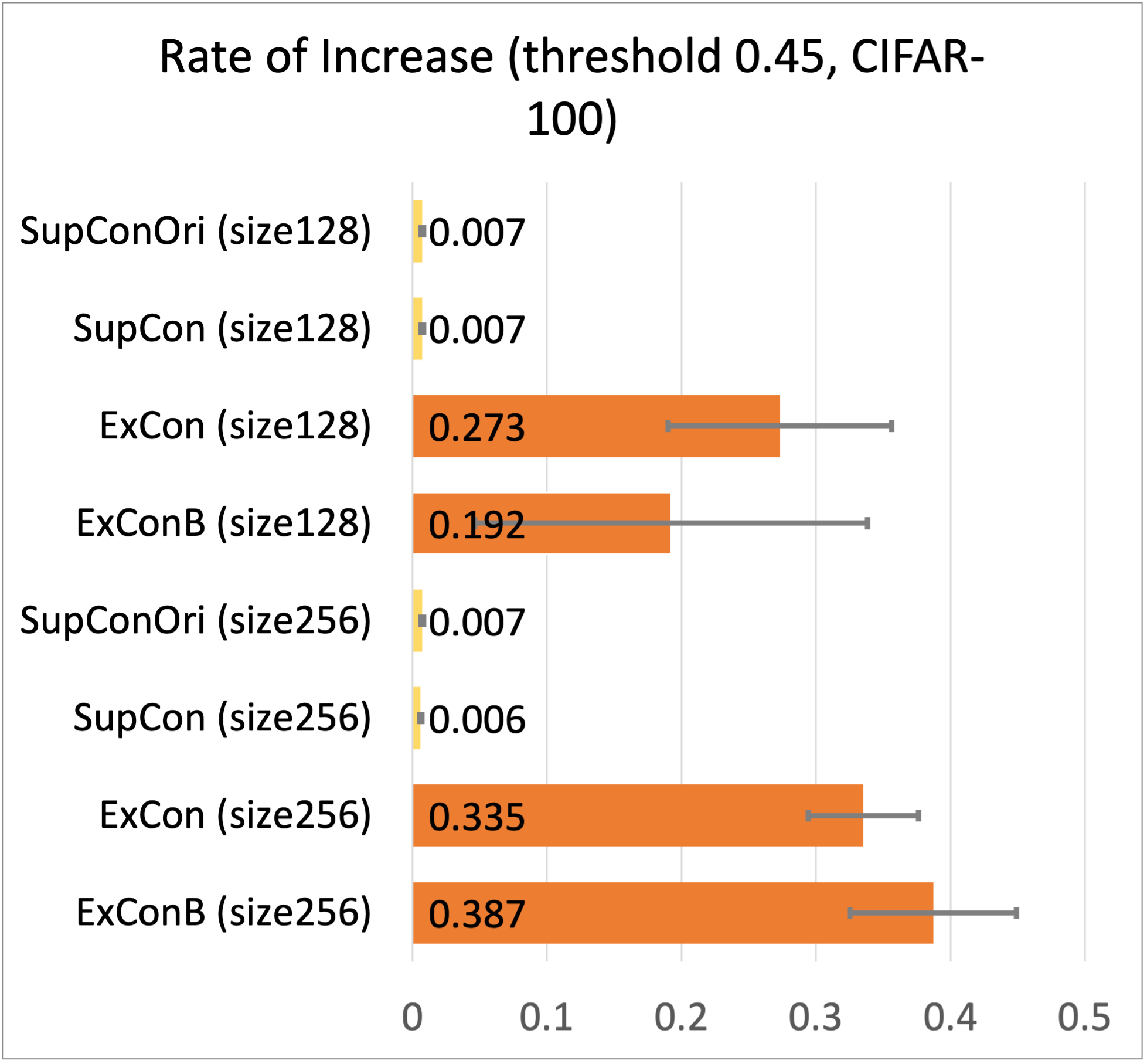}
    \caption{Average Drop \% (lower is better) and Rate of Increase in Scores (higher is better) of the models trained on CIFAR-100 for batch sizes of 128 and 256 respectively.}
    \label{fig:appendix_cifar_045}
\end{figure*}

\begin{figure*}
    \centering
    \includegraphics[width=0.55\textwidth]{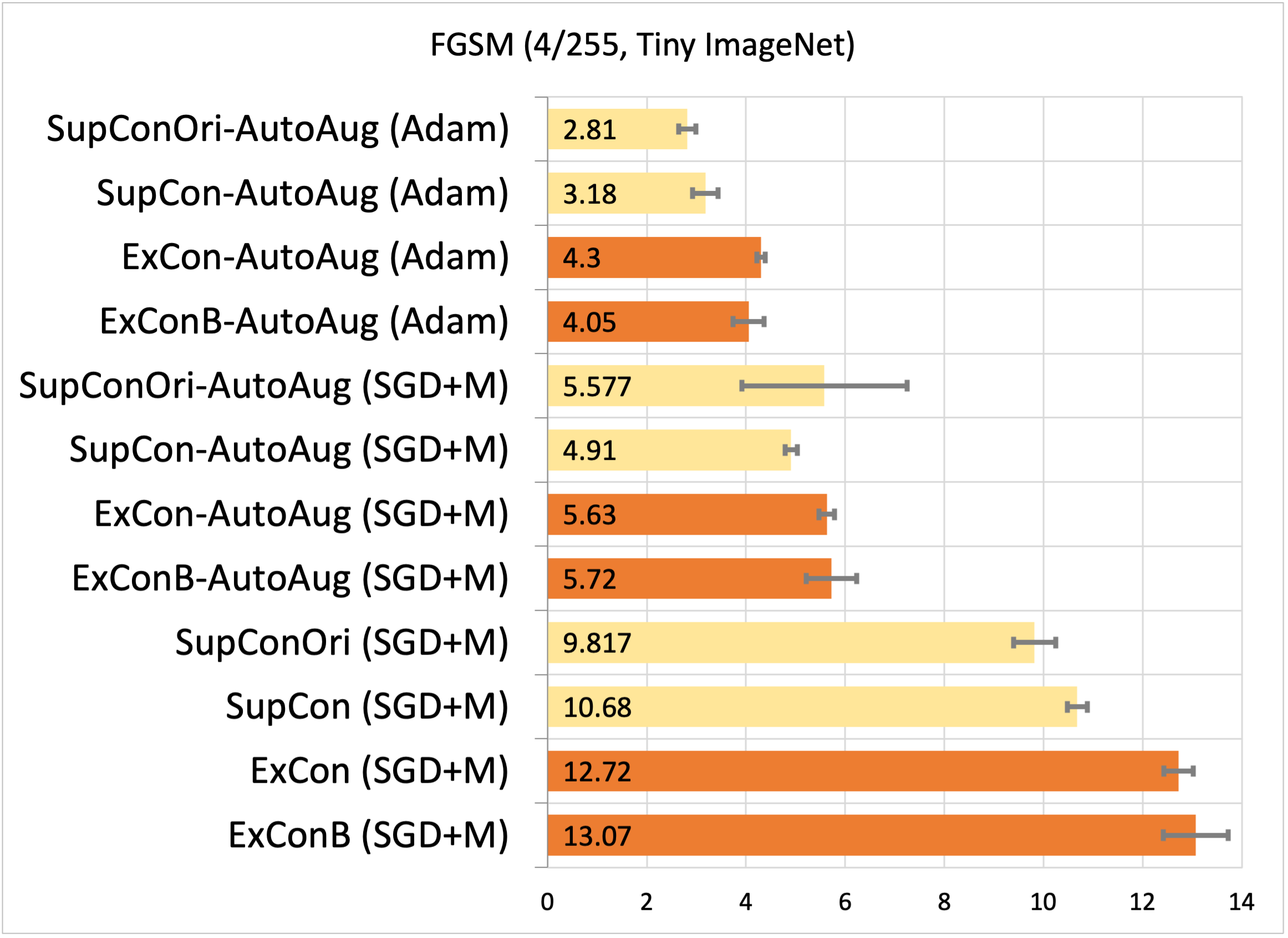}
    \includegraphics[width=0.43\textwidth]{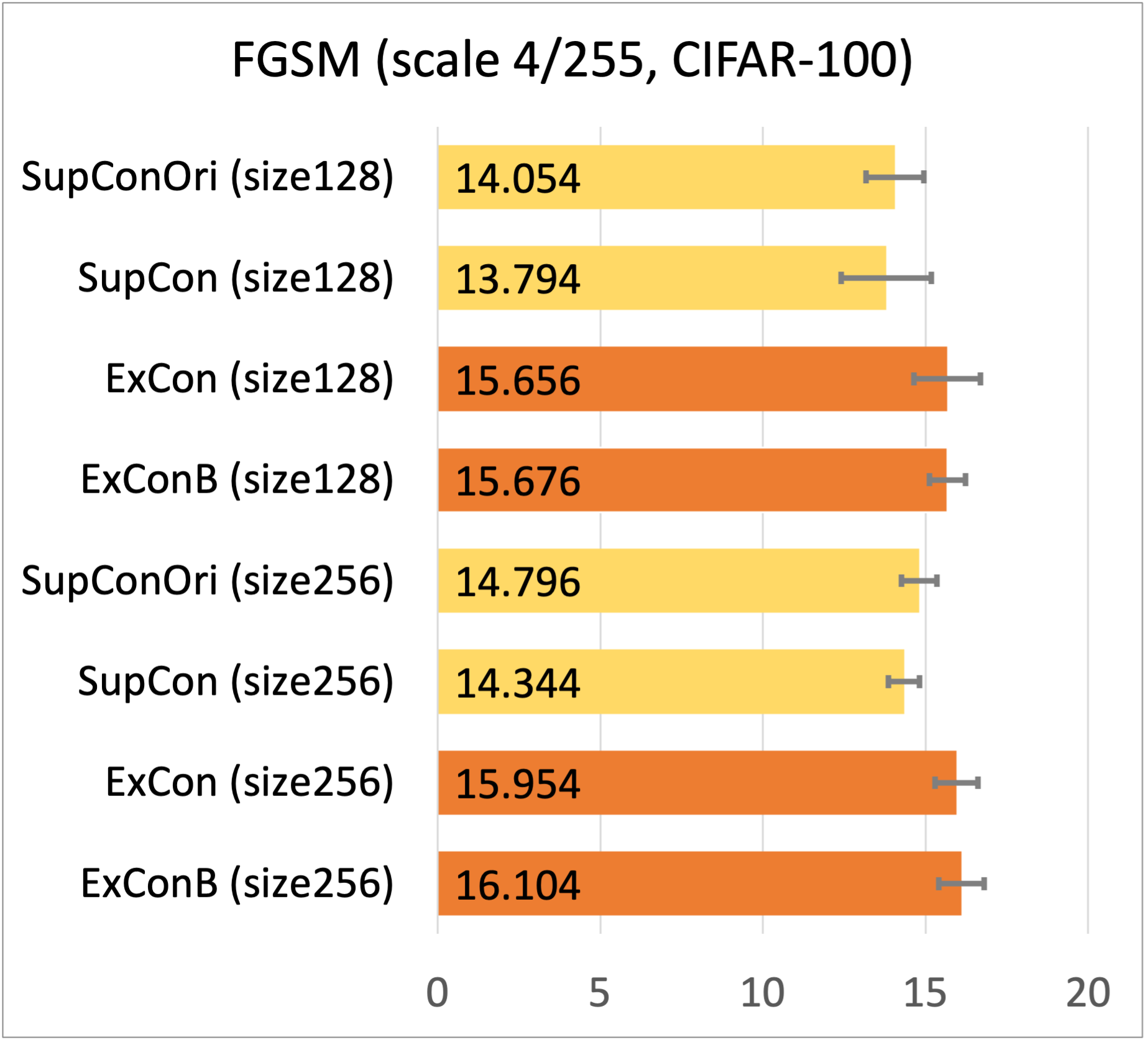}
    \caption{The top-1 accuracy under the FGSM input perturbation with the scale coefficient 4/255 for the Tiny ImageNet dataset and the CIFAR-100 dataset respectively. }
    \label{fig:appendix_fgsm}
\end{figure*}

\begin{figure*}[h]
    \centering
    \includegraphics[width=0.51\textwidth]{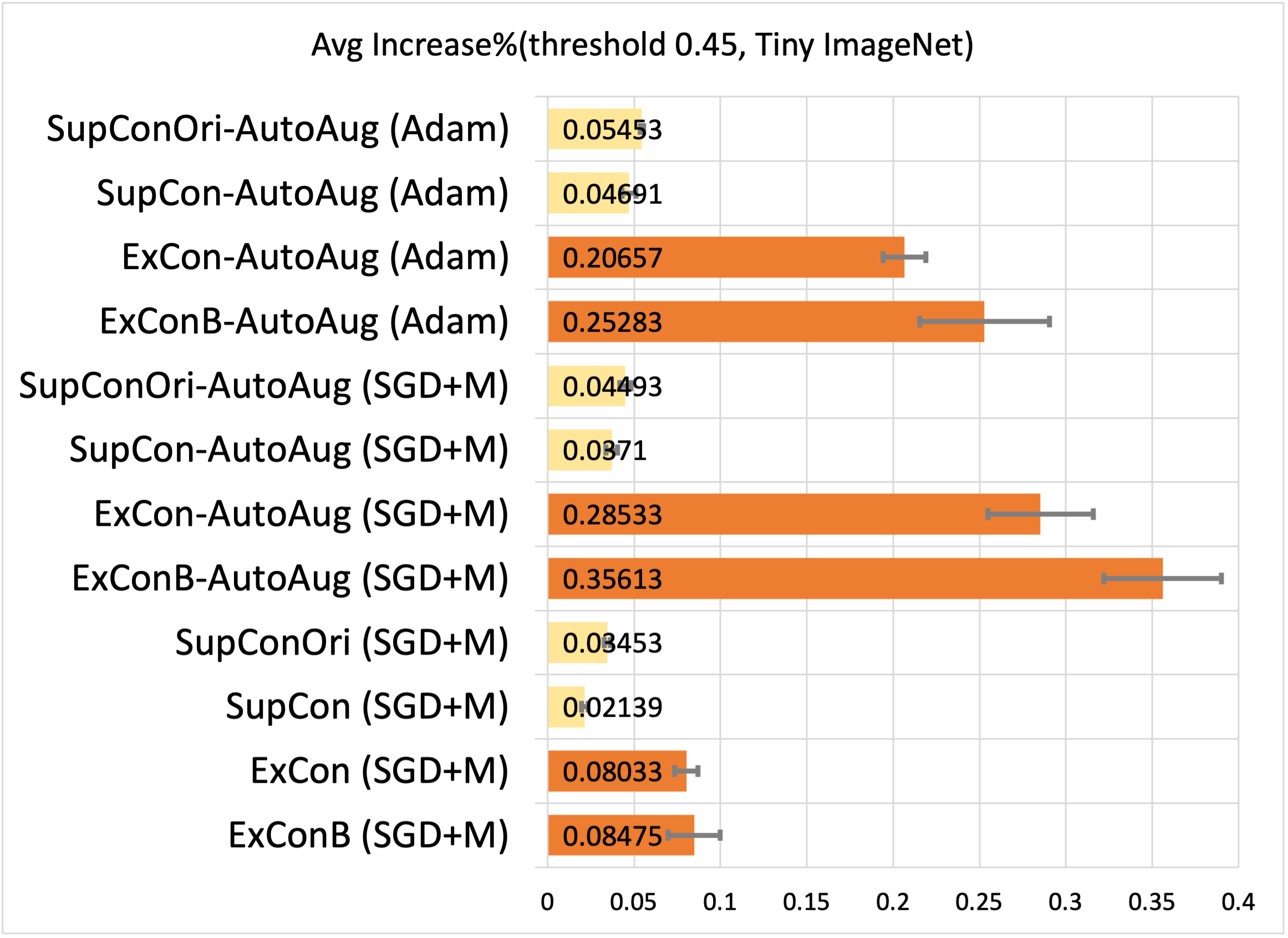}
    \includegraphics[width=0.45\textwidth]{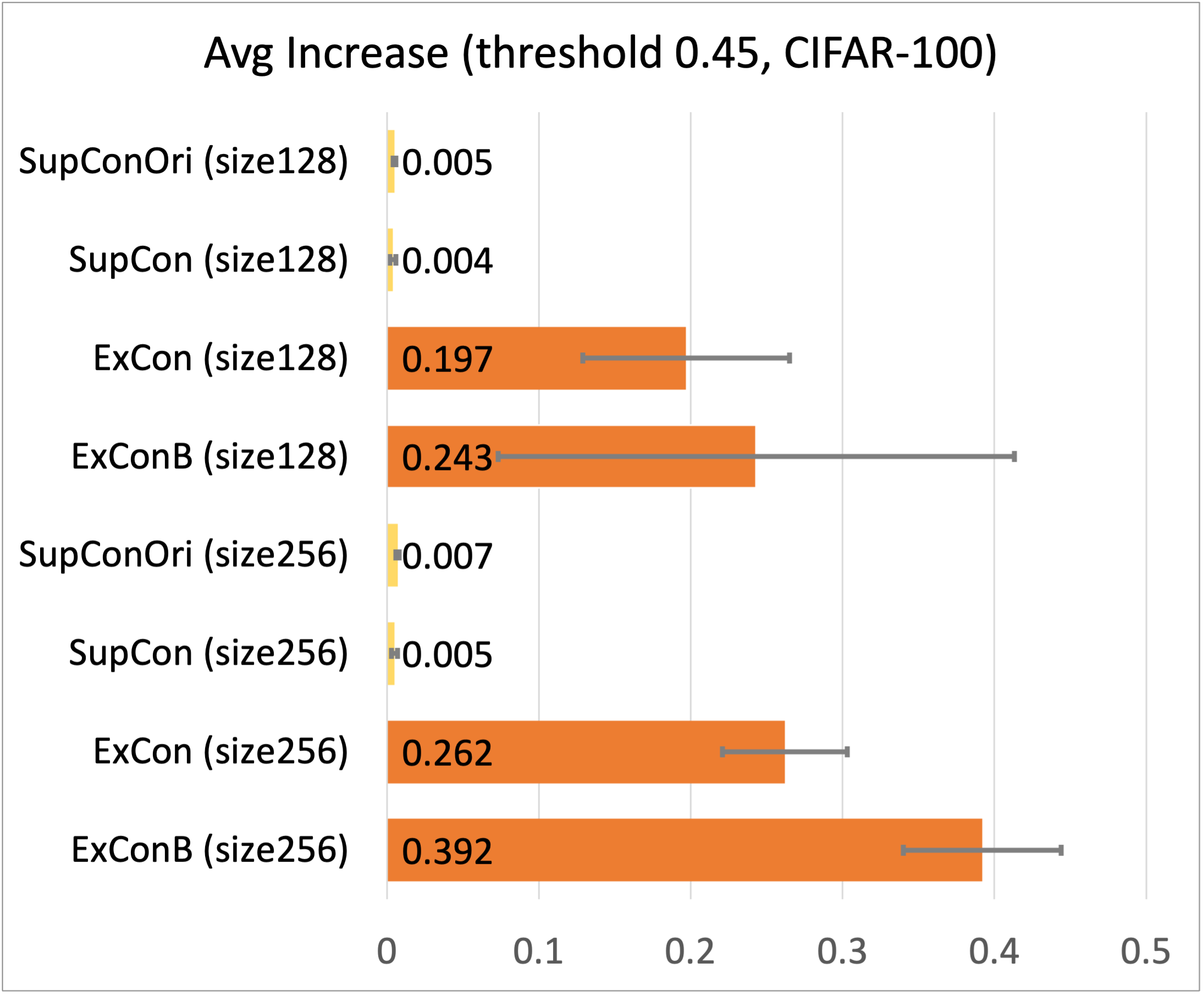}
    \caption{Average Increase \% (higher is better) where the top 45\% pixels are reserved.}
    \label{fig:appendix_045_avg_increase}
\end{figure*}

\begin{figure*}[h]
    \centering
    \includegraphics[width=0.51\textwidth]{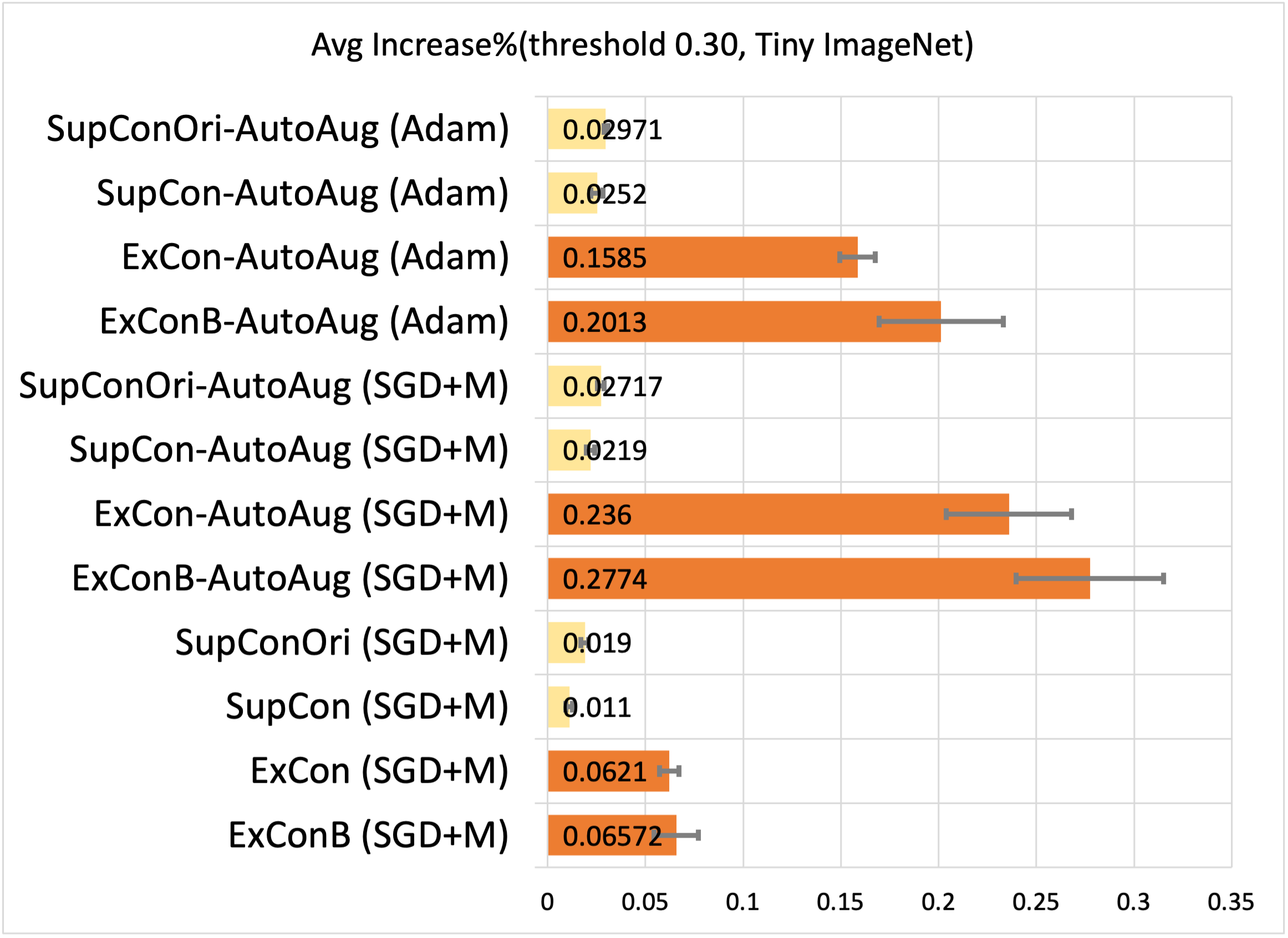}
    \includegraphics[width=0.45\textwidth]{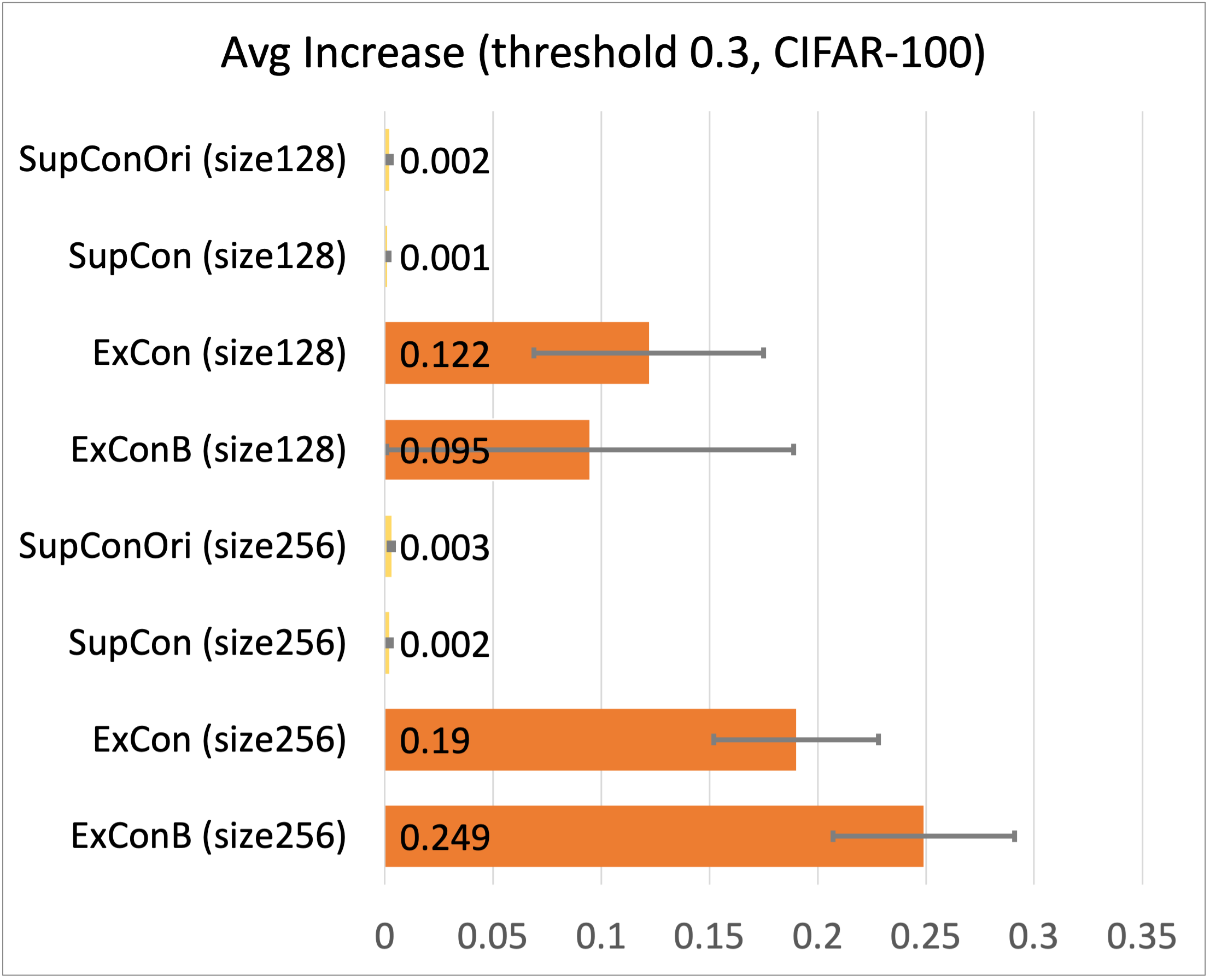}
    \caption{Average Increase \% (higher is better) where the top 30\% pixels are reserved.}
    \label{fig:appendix_030_avg_increase}
\end{figure*}

\begin{figure*}[h]
    \centering
    \includegraphics[width=0.51\textwidth]{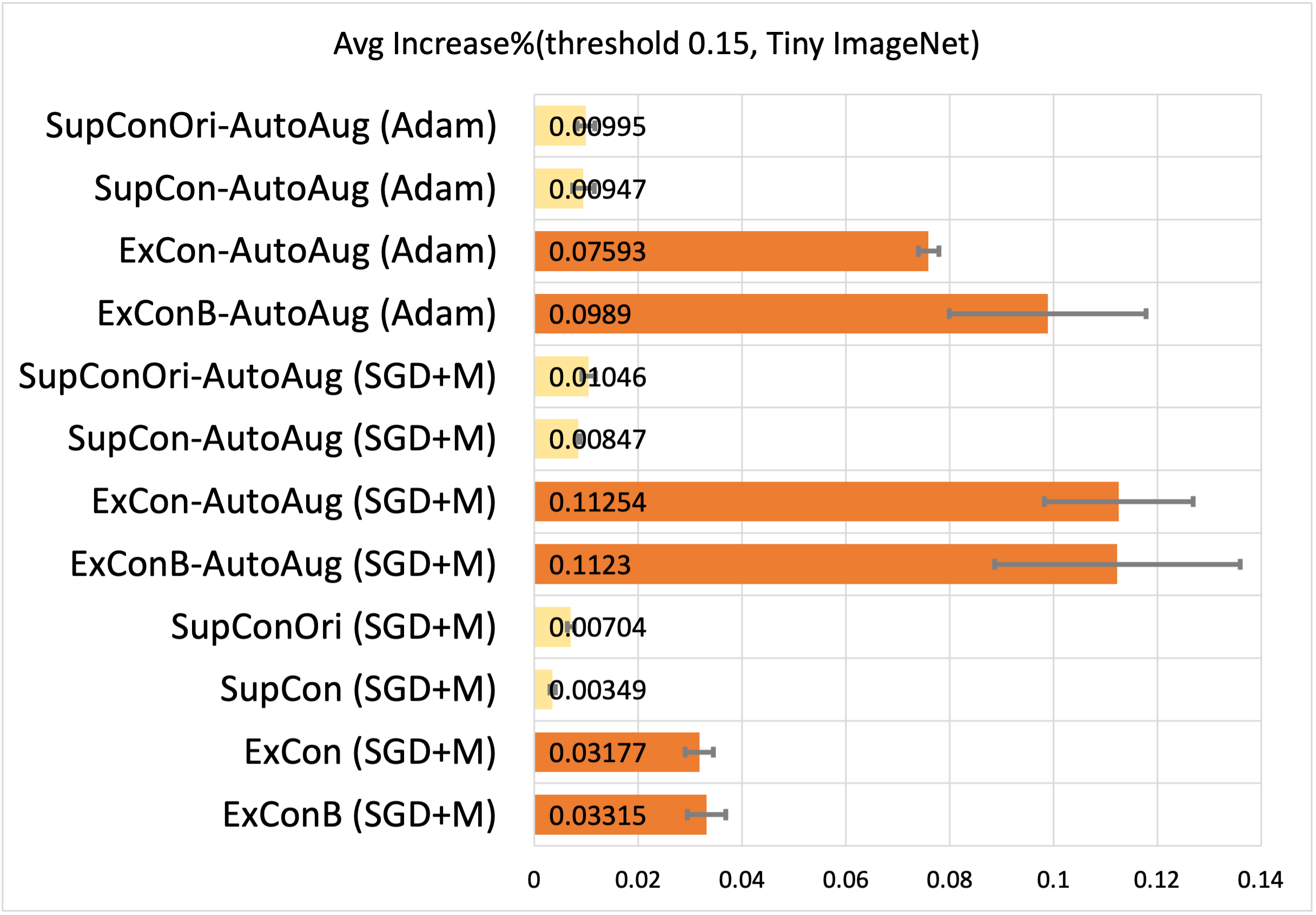}
    \includegraphics[width=0.45\textwidth]{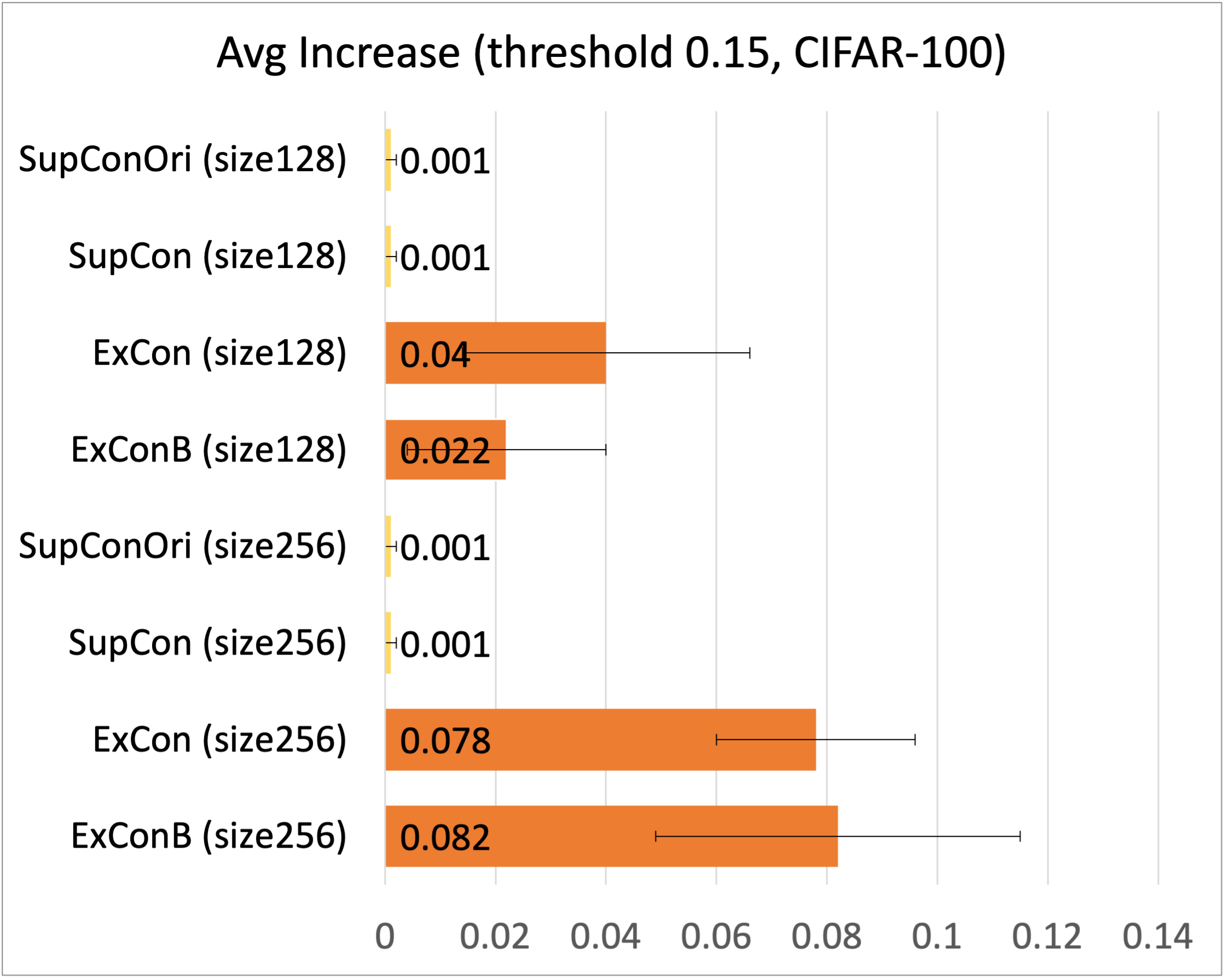}
    \caption{Average Increase \% (higher is better) where the top 15\% pixels are reserved.}
    \label{fig:appendix_015_avg_increase}
\end{figure*}

\begin{figure*}[h]
    \centering
    \includegraphics[width=0.51\textwidth]{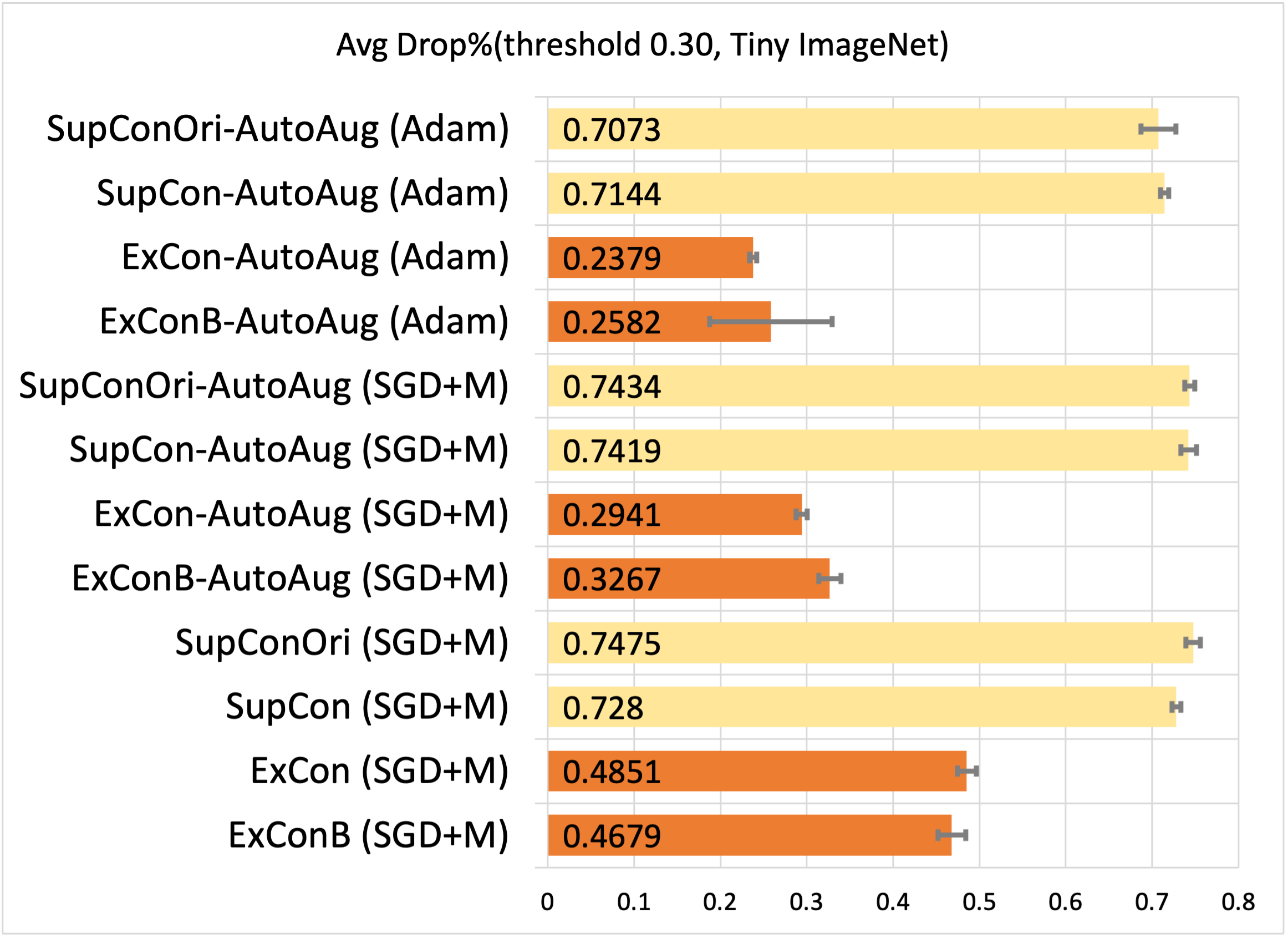}
    \includegraphics[width=0.45\textwidth]{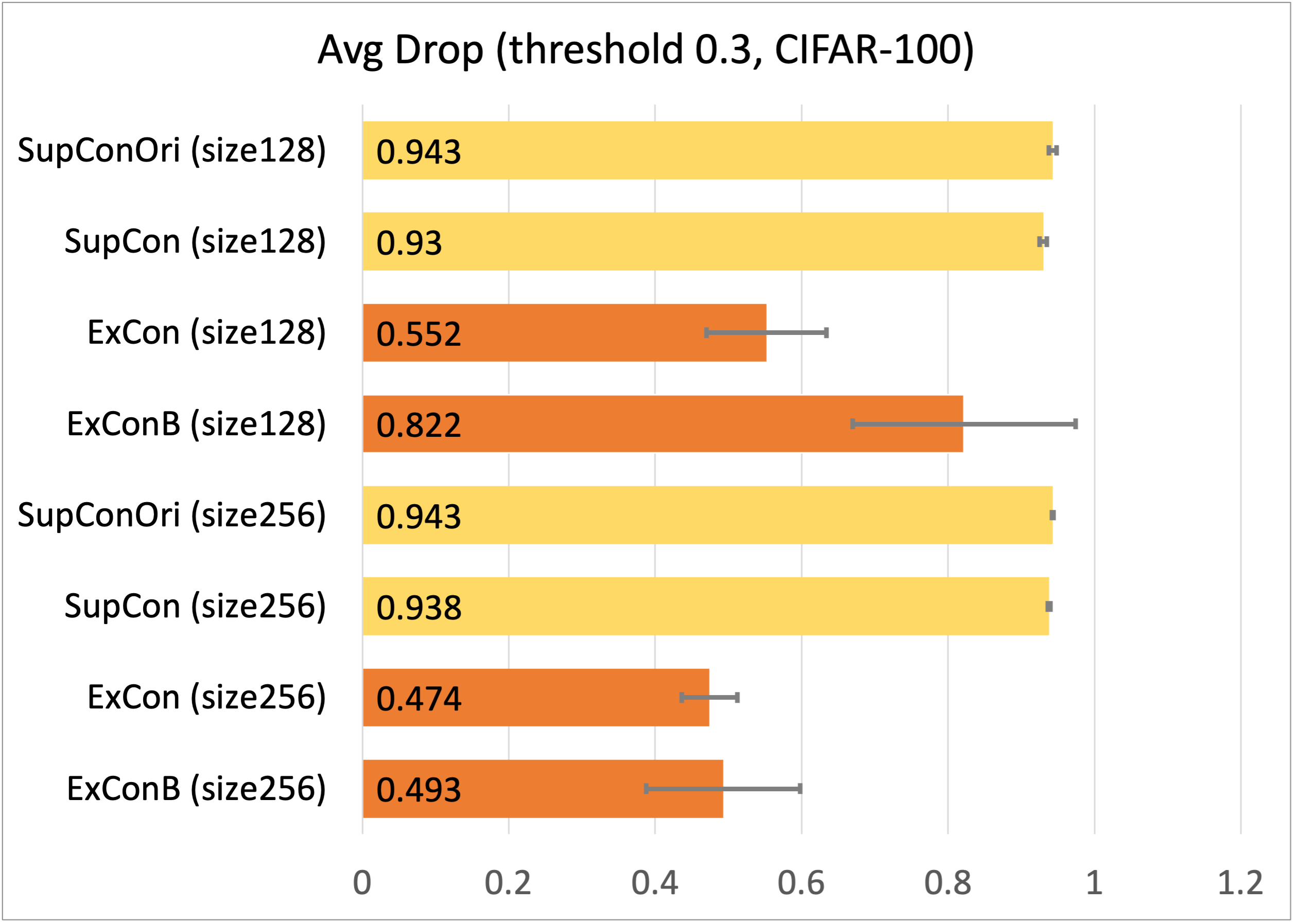}
    \caption{Average Drop \% (lower is better) where the top 30\% pixels are reserved.}
    \label{fig:appendix_030_avg_drop}
\end{figure*}

\begin{figure*}[h]
    \centering
    \includegraphics[width=0.51\textwidth]{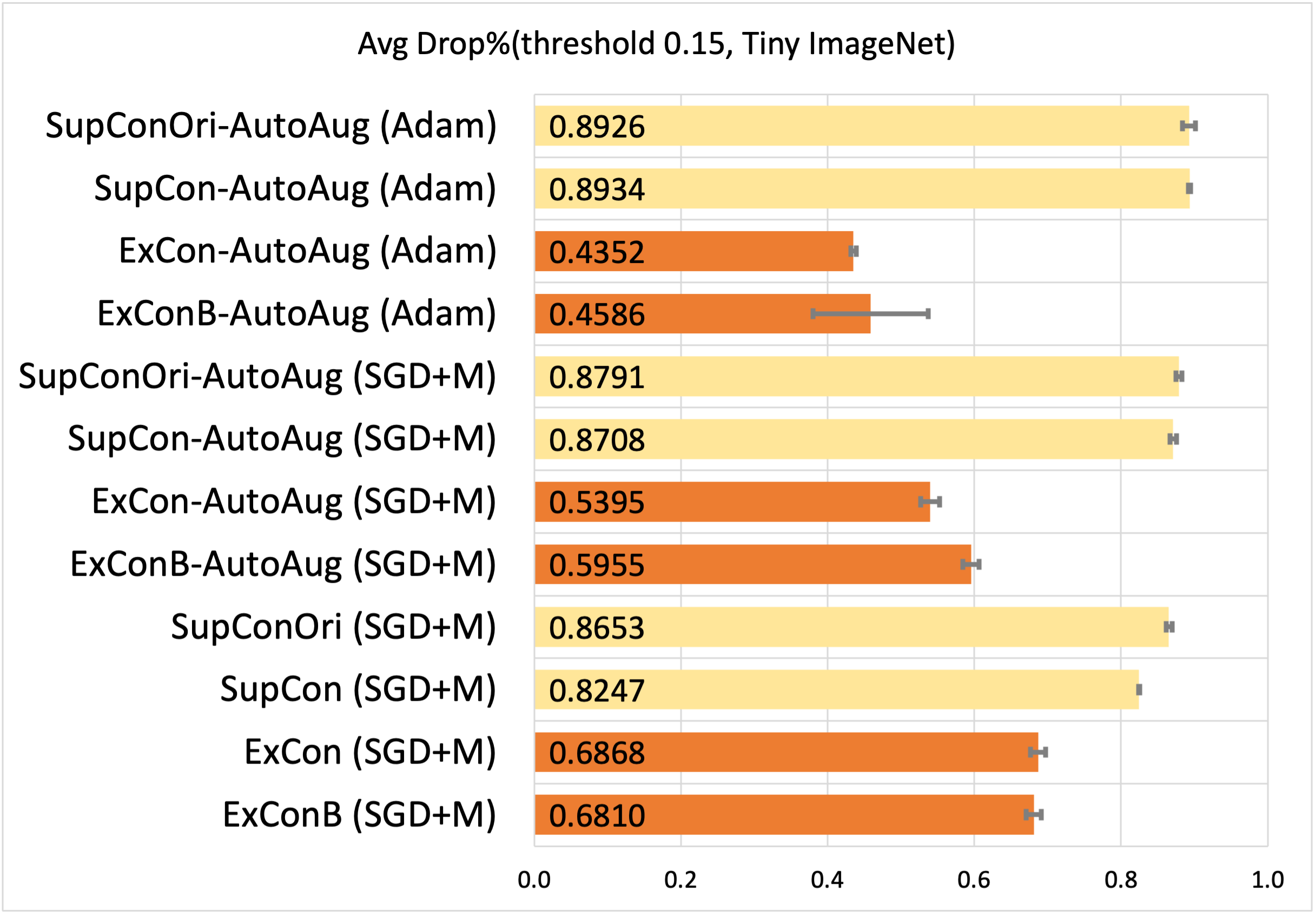}
    \includegraphics[width=0.45\textwidth]{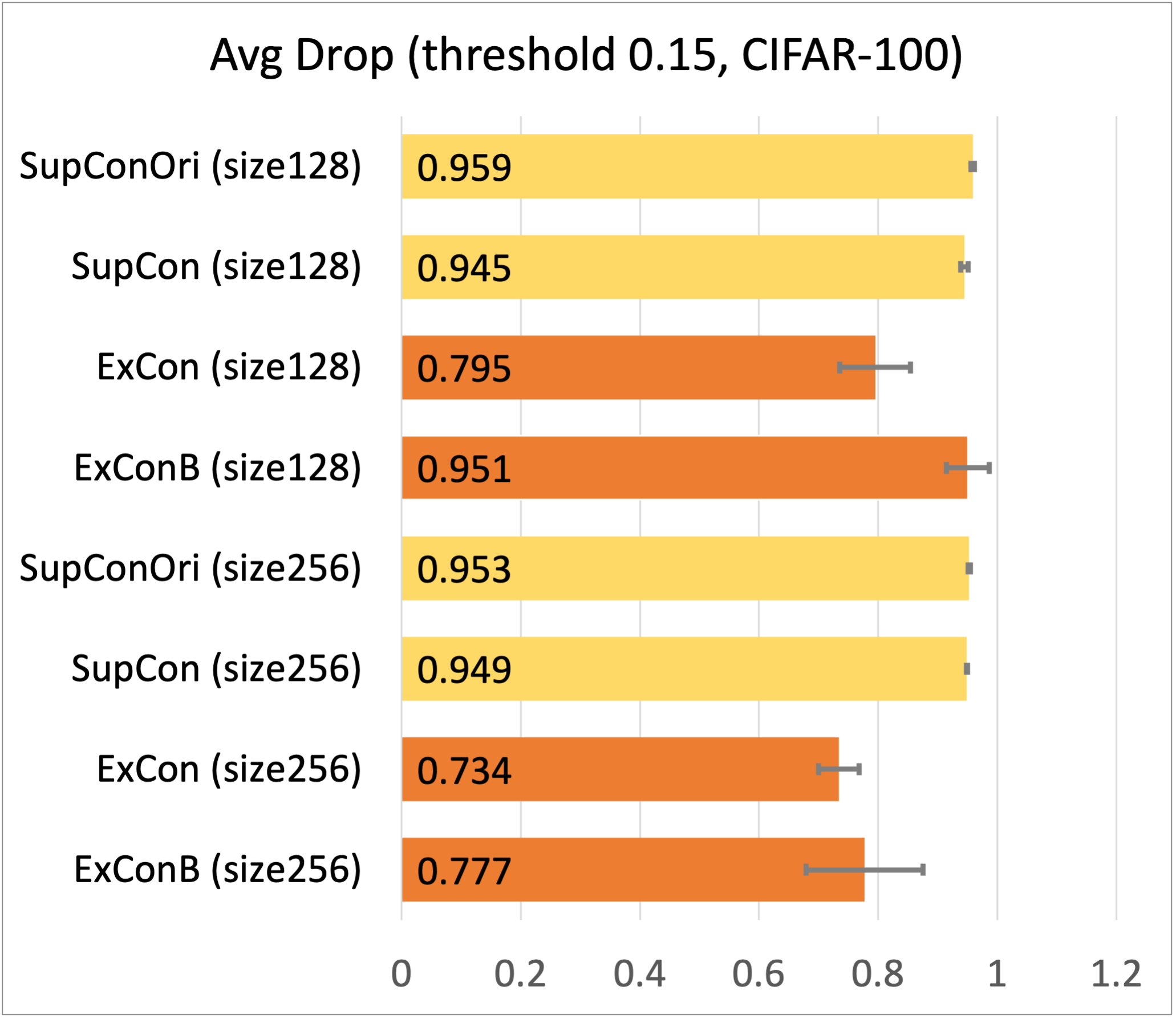}
    \caption{Average Drop \% (lower is better) where the top 15\% pixels are reserved.}
    \label{fig:appendix_015_avg_drop}
\end{figure*}

\begin{figure*}[h]
    \centering
    \includegraphics[width=0.51\textwidth]{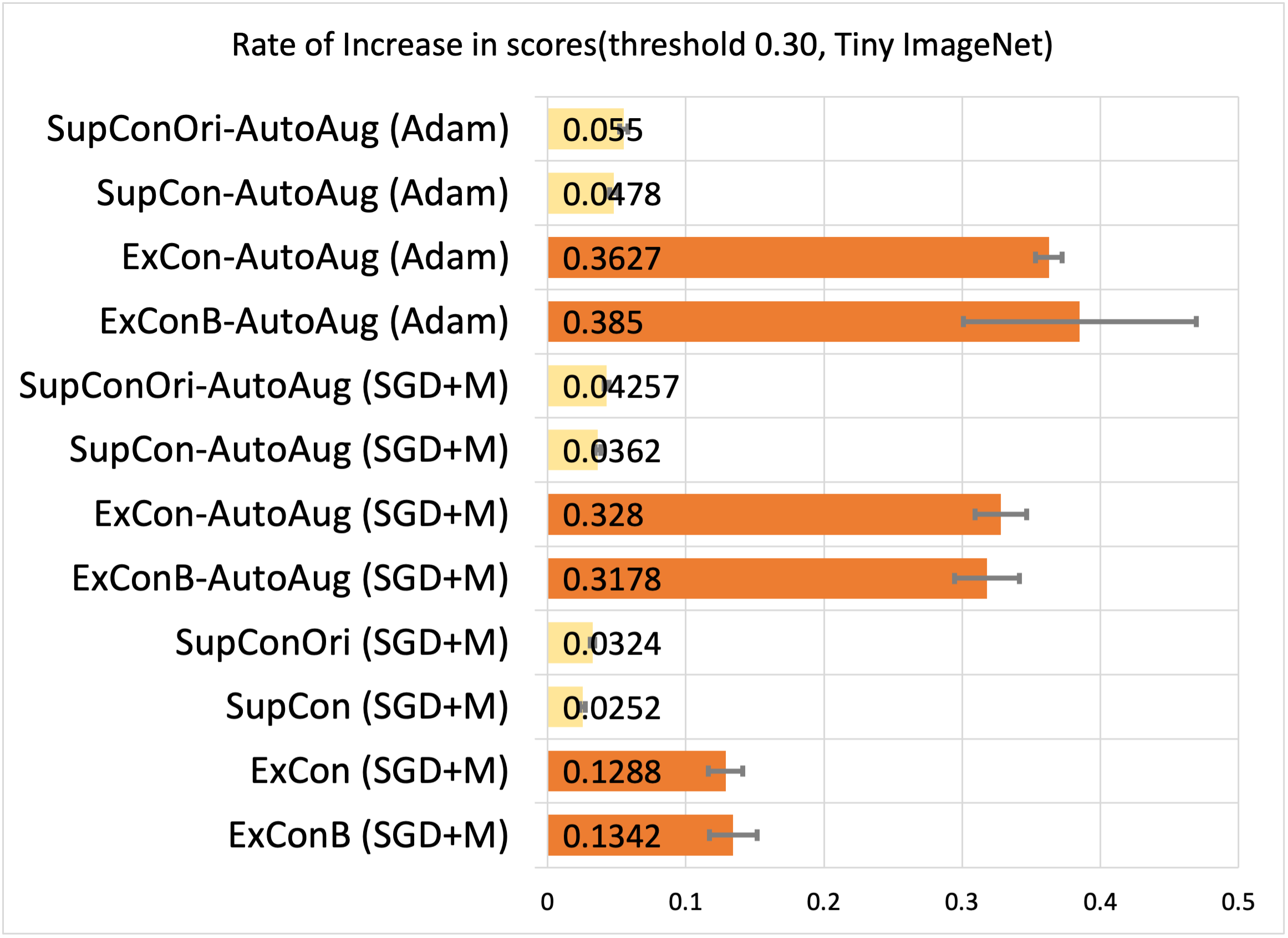}
    \includegraphics[width=0.45\textwidth]{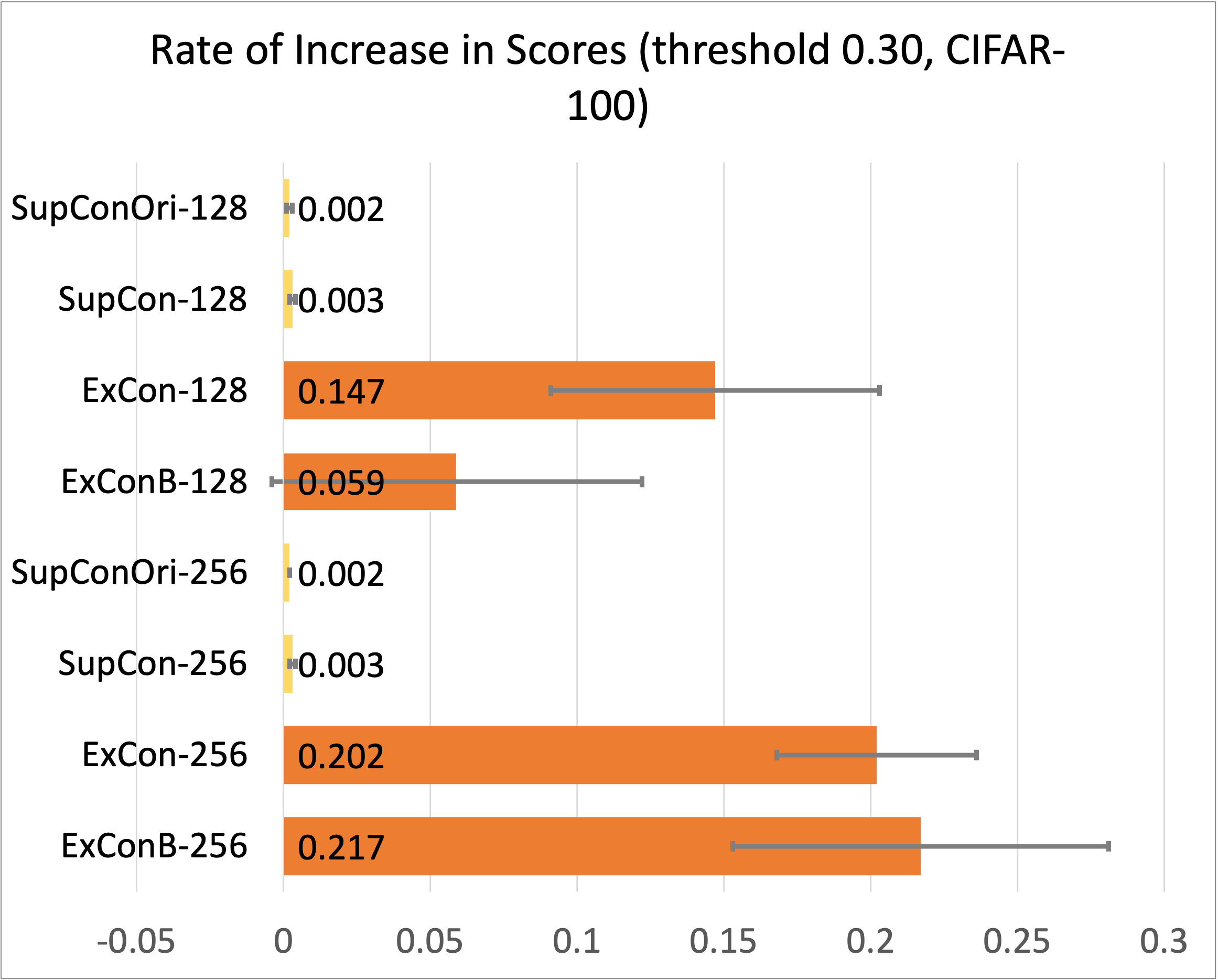}
    \caption{Rate of Increase in Scores (higher is better) where the top 30\% pixels are reserved.}
    \label{fig:appendix_030_rate_increase}
\end{figure*}

\begin{figure*}[h]
    \centering
    \includegraphics[width=0.51\textwidth]{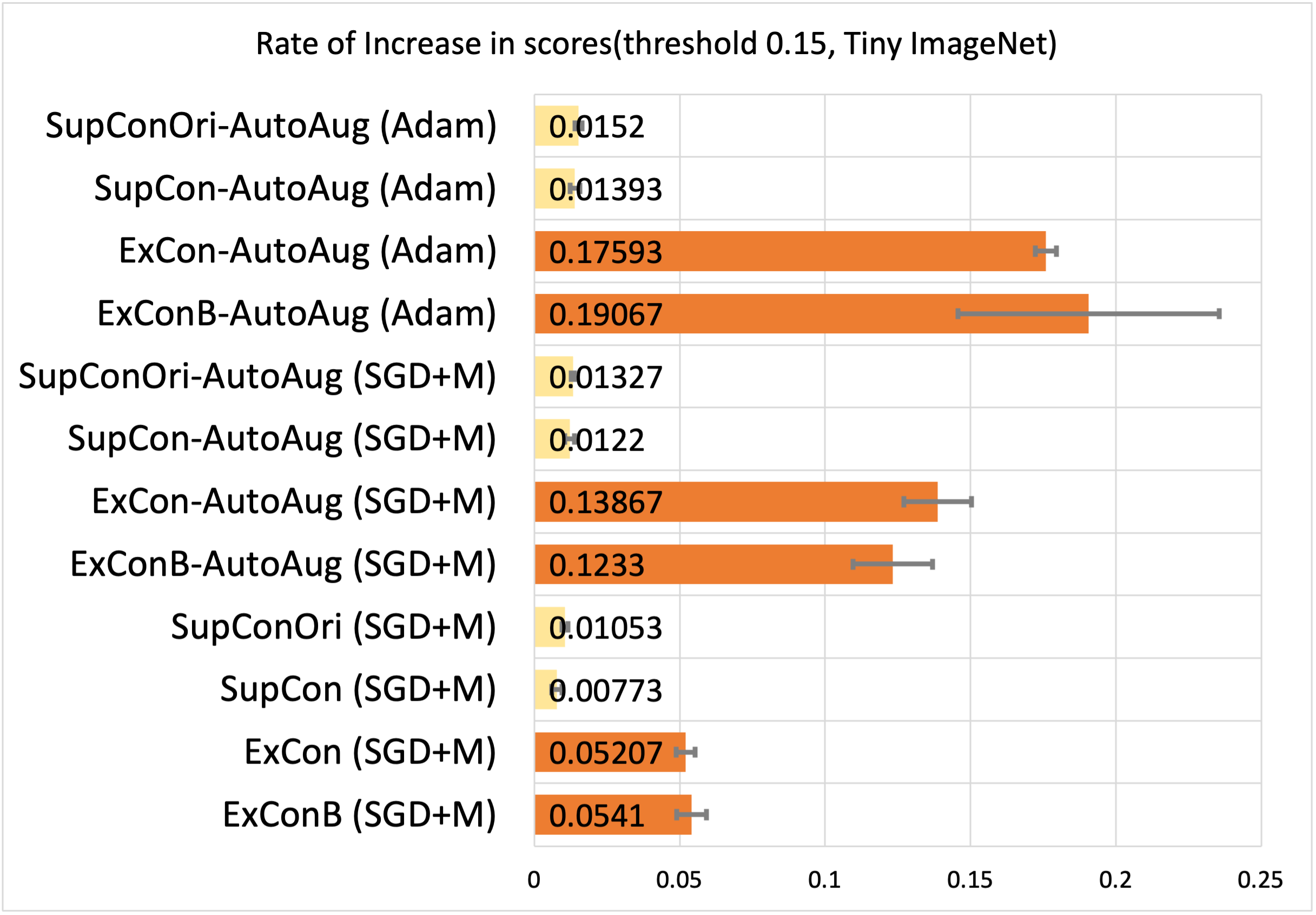}
    \includegraphics[width=0.45\textwidth]{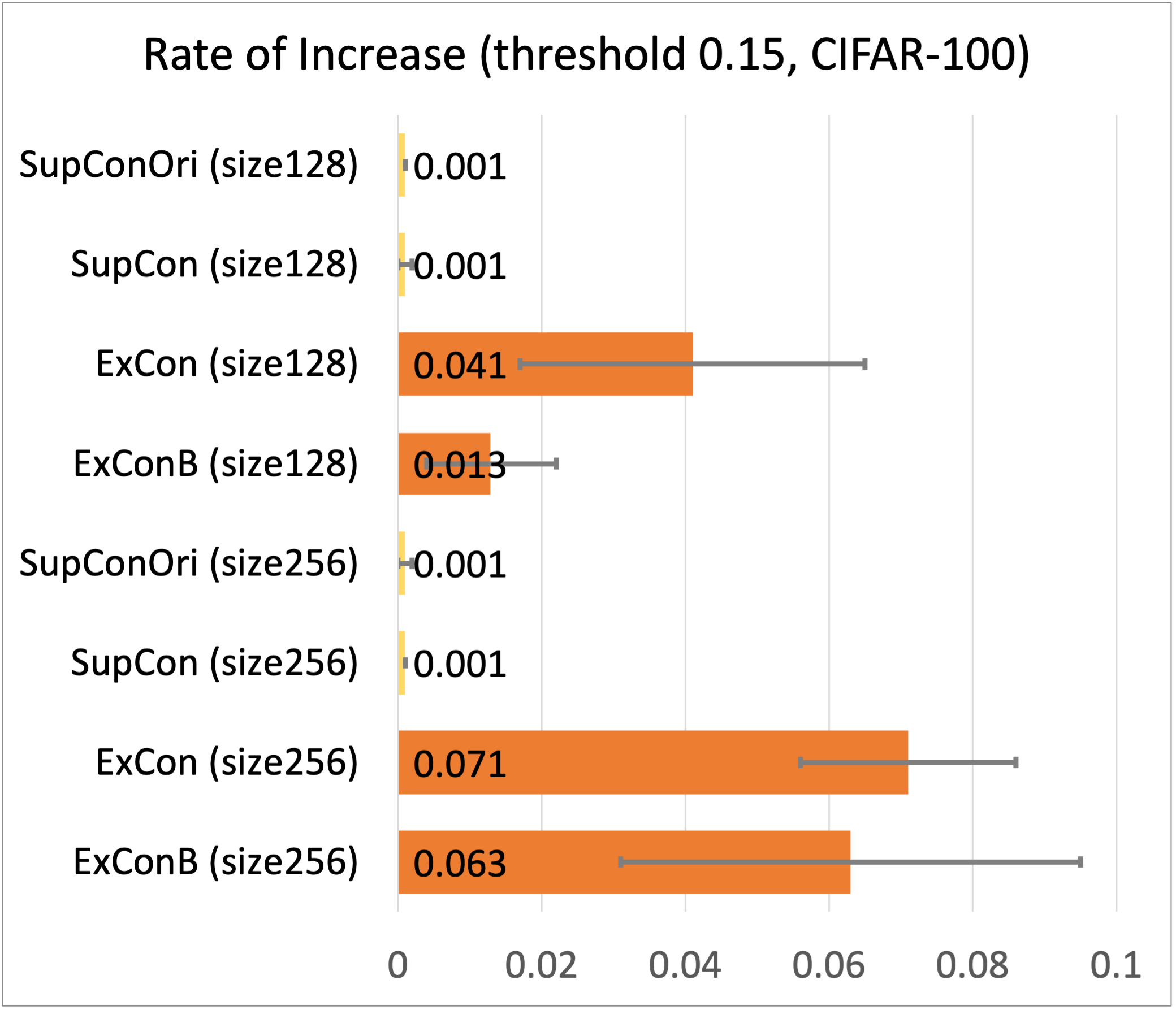}
    \caption{Rate of Increase in Scores (higher is better) where the top 15\% pixels are reserved.}
    \label{fig:appendix_015_rate_increase}
\end{figure*}

\begin{figure*}
    \centering

        \includegraphics[width=0.49\textwidth]{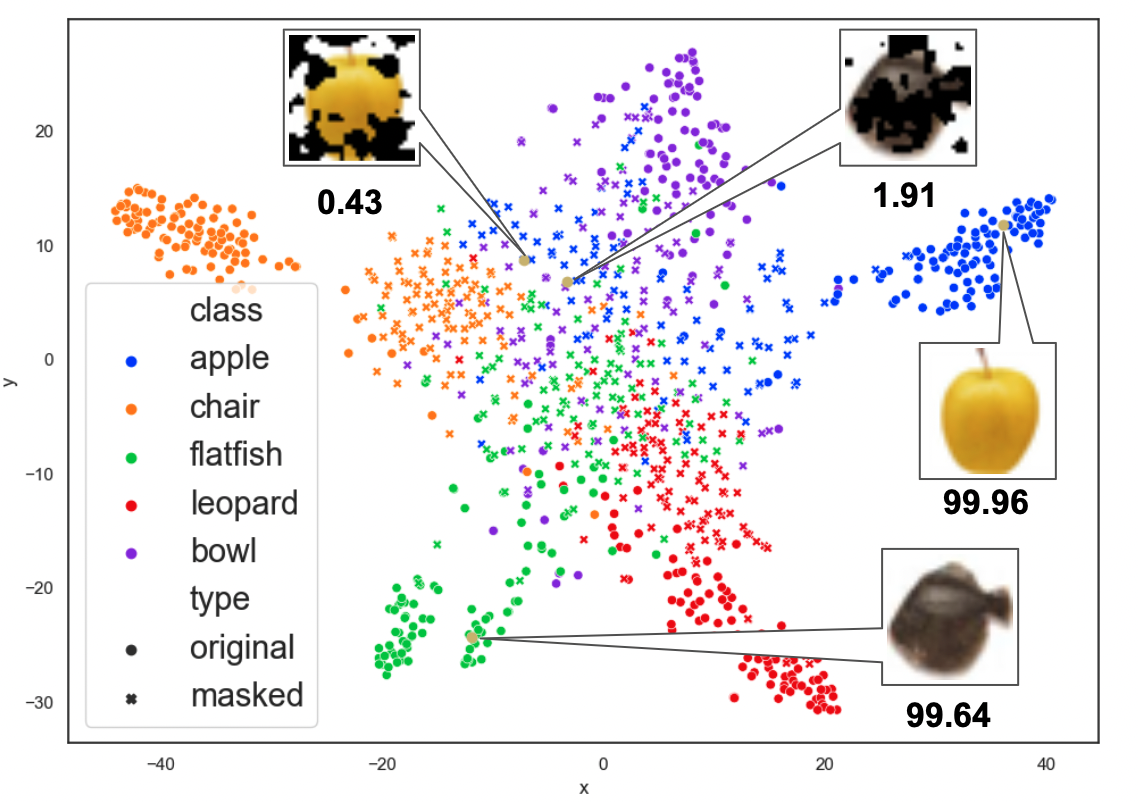}
        \includegraphics[width=0.49\textwidth]{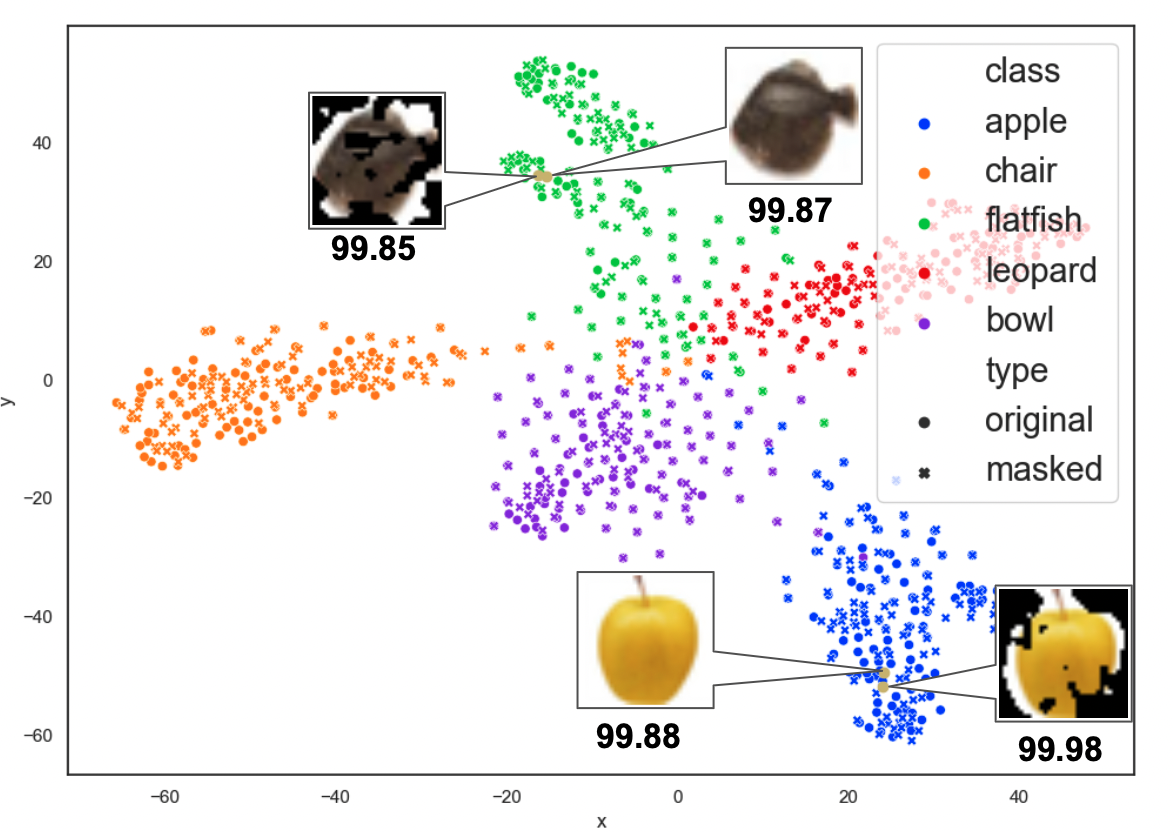}

    \caption{t-SNE embeddings for SupConOri (left) and ExCon (right) on the CIFAR-100 dataset.}
    \label{fig:embedding_cifar100}
\end{figure*}

\end{document}